\documentclass[10pt,twocolumn,letterpaper]{article}

\usepackage{cvpr}              
\usepackage{graphicx}
\usepackage{amssymb}
\usepackage{amsfonts}
\usepackage{amsmath}
\usepackage{bm}
\usepackage{mathrsfs}
\usepackage{makecell}
\usepackage{multirow}
\usepackage{multicol}
\usepackage{booktabs}
\usepackage{cuted}
\usepackage{capt-of}
\usepackage{placeins}
\usepackage{stackengine} 
\usepackage{indentfirst}
\usepackage{algorithm}
\usepackage[noend]{algpseudocode}
\usepackage{tabularray}
\usepackage{setspace}
\usepackage[accsupp]{axessibility}

\RequirePackage{tabularray}
\usepackage[table]{xcolor}

\usepackage{color}
\definecolor{darkgreen}{rgb}{0,0.5,0.7}
\definecolor{lightgreen}{rgb}{0.4,0.8,0.4}
\definecolor{lightorange}{rgb}{0.9,0.6,0.1}
\definecolor{darkpurple}{rgb}{0.5,0,0.7}
\definecolor{darkblue}{rgb}{0.5,0,0.7}
\definecolor{Comment}{rgb}{0.62,0.478,0.259}

\definecolor{RayRefinement}{rgb}{0.494,0.392,0.620}
\definecolor{Fthetal}{rgb}{0.573,0.224,0.196}
\definecolor{Sgtq}{rgb}{0.553,0.486,0.357}
\definecolor{St}{rgb}{0.749,0.565,0}
\newcommand{\draft}[2]{\textcolor{#1}{#2}}

\newcommand{\best}[1] {\textbf{\draft{red}{#1}}}
\newcommand{\second}[1] {\underline{\draft{blue}{#1}}}

\definecolor{process}{rgb}{0.216,0.537,0.620}
\newcommand{\process}[1] {\textcolor{process}{#1}}
\newcommand{\load}[1] {\textcolor{lightgreen}{#1}}
\newcommand{\unload}[1] {\textcolor{lightorange}{#1}}
\newcommand{\train}[1] {\textcolor{Fthetal}{#1}}
\newcommand{\frozen}[1] {\textcolor{darkblue}{#1}}

\usepackage[
    enable,
    labname={CMLab},
    logo={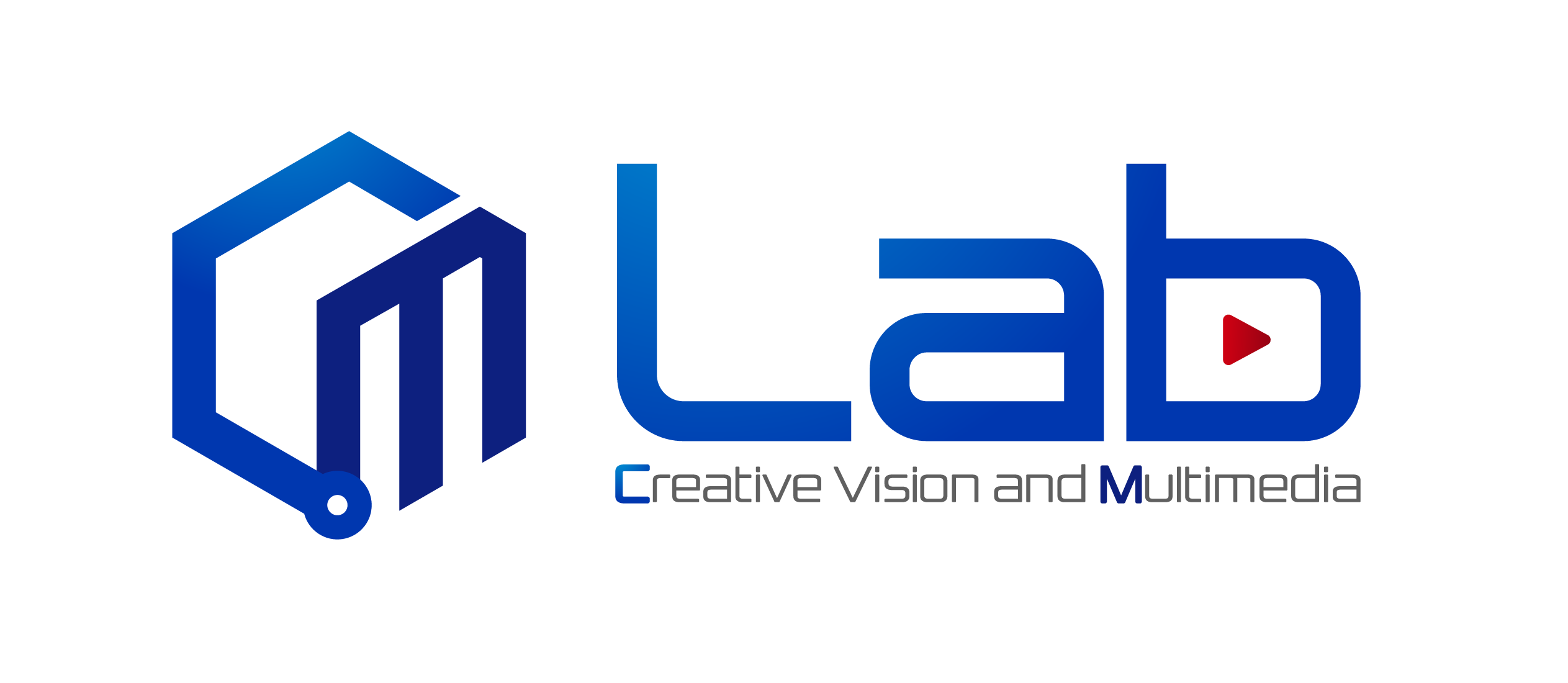},
    leftlogo={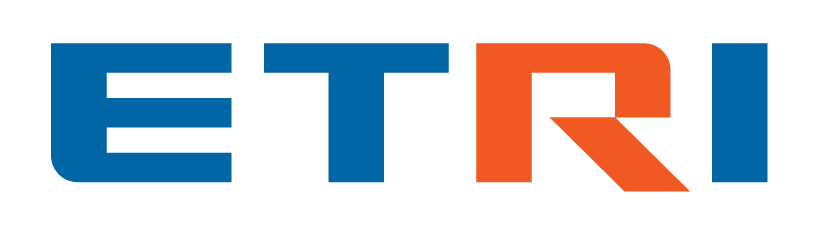},
    titleicon={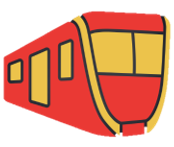},
    logowidth={22mm},
    projectpage={https://cmlab-korea.github.io/MoRel/}
]{cmlab}
\definecolor{cvprblue}{rgb}{0.21,0.49,0.74}
\usepackage[pagebackref,breaklinks,colorlinks,allcolors=cvprblue]{hyperref}

\def\paperID{19079} 
\def\confName{CVPR}
\def\confYear{2026}
\begin{document}

\title{MoRel: Long-Range Flicker-Free 4D Motion Modeling via Anchor Relay-based Bidirectional Blending with Hierarchical Densification}
\vspace{-5mm}


\cmlabAuthors{\text{Sangwoon Kwak}$^{1}$\footnotemark[1]\hspace{2em} \text{Weeyoung Kwon}$^{2}$\footnotemark[1]\hspace{2em}  \text{Jun Young Jeong}$^{1}$\hspace{2em} \text{Geonho Kim}$^{2}$ \\ \text{Won-Sik Cheong}$^{1}$\hspace{2em}\quad \text{Jihyong Oh}$^{2}$\footnotemark[2]\\[0.5em]}

\cmlabAuthorsFootnotes{$^{*}$ Equal contribution \qquad $^{\dagger}$ Corresponding author}
\cmlabAffiliations{$^{1}$ ETRI \qquad$^{2}$ CMLab, Chung-Ang University}
\cmlabAuthorEmail{$\{$s.kwak, jyj0120, wscheong$\}$@etri.re.kr, $\{$weeyoungkwon, joelkimgh, jihyongoh$\}$@cau.ac.kr}
\cmlabProjectPage{https://cmlab-korea.github.io/MoRel/}
\maketitle
\refstepcounter{figure}\label{fig:figure_page1}
\begin{strip}
\vspace{-8mm}
\centering
\includegraphics[width=\textwidth]{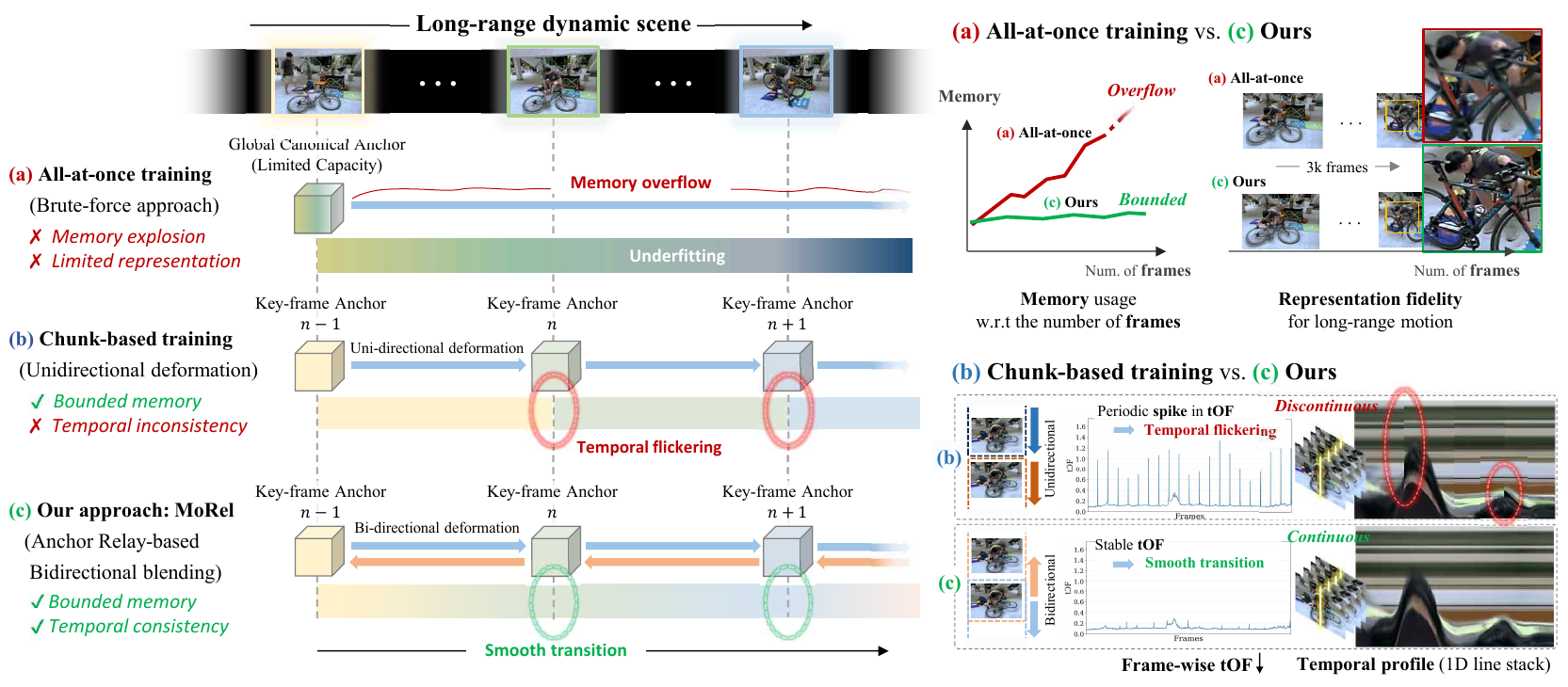}

\par\vspace{1mm}
\parbox{0.98\textwidth}{\small
Figure \thefigure. \textbf{Approaches for modeling long-range 4D Motion.}
(a) The all-at-once training experiences memory overflow and even suffers from limited representational capacity.
(b) The chunk-based training mitigates the memory overflow but causes temporal flickering at chunk boundaries, substantially degrading visual quality.
In contrast, (c) our Anchor Relay-based Bidirectional Blending (ARBB) approach successfully maintains both representation quality and temporal consistency by smoothly transitioning the influence of each Key-frame Anchor (KfA).
The rendered patches, frame-wise tOF~\cite{chu2018temporally}, and temporal profile provide strong evidence for the effectiveness of our method.
}
\vspace{-4mm}
\end{strip}

{
  \renewcommand{\thefootnote}%
    {\fnsymbol{footnote}}
    \footnotetext[1]{Equal contribution.}
  \footnotetext[2]{Corresponding author.}
}
\begin{abstract}
Recent advances in 4D Gaussian Splatting (4DGS) have extended the high-speed rendering capability of 3D Gaussian Splatting (3DGS) into the temporal domain, enabling real-time rendering of dynamic scenes.
However, one of the major remaining challenges lies in modeling long-range motion-contained dynamic videos, where a naïve extension of existing methods leads to severe memory explosion, temporal flickering, and failure to handle appearing or disappearing occlusions over time. 
To address these challenges, we propose a novel 4DGS framework characterized by an Anchor Relay-based Bidirectional Blending (ARBB) mechanism, named MoRel, which enables temporally consistent and memory-efficient modeling of long-range dynamic scenes.
Our method progressively constructs locally canonical anchor spaces at key-frame time indices and models inter-frame deformations at the anchor level, enhancing temporal coherence. 
By learning bidirectional deformations between KfAs and adaptively blending them through learnable opacity control, our approach mitigates temporal discontinuities and flickering artifacts.
We further introduce a Feature-variance-guided Hierarchical Densification (FHD) scheme that effectively densifies KfA's while keeping rendering quality, based on an assigned level of feature-variance.
To effectively evaluate our model's capability to handle real-world long-range 4D motion, we newly compose a long-range 4D motion-contained dataset, called SelfCap$_{\text{LR}}$. It has larger average dynamic motion magnitude, captured at spatially wider spaces, compared to previous dynamic video datasets.
Overall, our MoRel achieves temporally coherent and flicker-free long-range 4D reconstruction while maintaining bounded memory usage, demonstrating both scalability and efficiency in dynamic Gaussian-based representations. The code and project page:~\href{https://cmlab-korea.github.io/MoRel/}{https://cmlab-korea.github.io/MoRel/}
\end{abstract} 
\vspace{-4mm}
\section{Introduction}
\label{sec:Introduction}

\begin{figure}[t]
\centering
\includegraphics[scale=0.53]{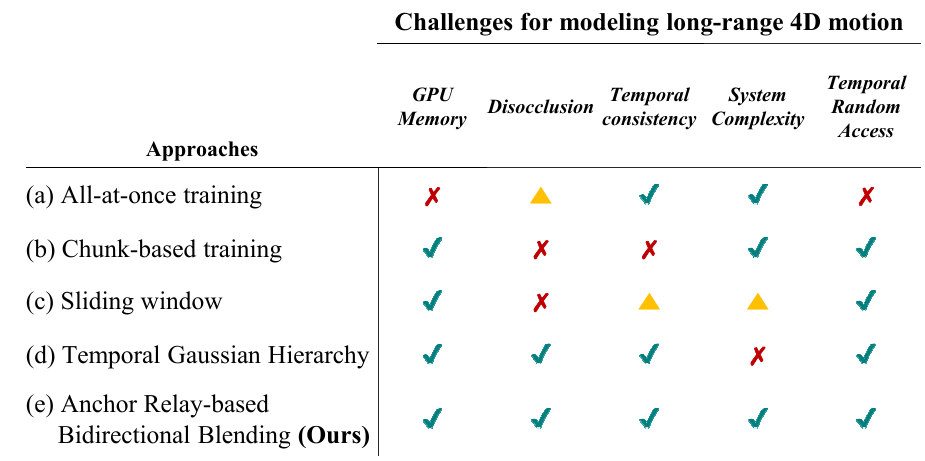}
\vspace{-2mm}
\caption{\textbf{Conceptual comparison of existing 4DGS methods in modeling long-range 4D motion.} (a) All-at-once approaches suffer from high memory usage, while (b) chunk-based methods inevitably fail to maintain temporal consistency. Even advanced variants struggle with system applicability such as a random accessibility. Our ARBB framework resolves all these issues, achieving bounded memory and temporally coherent long-range modeling.}
\label{fig:comparison_table}
\vspace{-6mm}
\end{figure}

3D Gaussian Splatting (3DGS)~\cite{kerbl20233d} has positioned itself as a powerful paradigm for novel view synthesis (NVS), enabling real-time photorealistic rendering. Unlike Neural Radiance Fields (NeRF)~\cite{mildenhall2021nerf}, which rely on dense ray sampling and costly volume rendering, 3DGS uses explicit Gaussian primitives and a GPU-parallel splatting pipeline to achieve high efficiency while preserving visual fidelity. This remarkable balance has naturally motivated its extension to dynamic and video-centric scenarios, giving rise to 4D Gaussian Splatting (4DGS)~\cite{wu20244d, yang2023gs4d, shaw2024swings, xu2024longvolcap}. However, existing 4DGS approaches encounter distinct challenges when modeling long-range videos, even those lasting only a few minutes, which are typical of real-world content. As summarized in Fig.~\ref{fig:comparison_table}, these methods involve clear trade-offs across different practical dimensions, highlighting the need for more scalable and versatile 4DGS solutions.
    
The all-at-once training approach (Fig.~\ref{fig:figure_page1}(a), Fig.~\ref{fig:comparison_table}(a)) extends existing methods to long-range videos by either (i) augmenting a canonical 3DGS with a deformation field ~\cite{huang2023sc, deformgs, wang2025freetimegs, li2023spacetime, wu2025swift4d} or (ii) using 4D Gaussian primitives that jointly encode spatial-temporal characteristics ~\cite{wu20244d, duan20244d, yang2023gs4d}. While jointly optimizing all frames ensures global temporal consistency, it causes GPU memory explosion because modeling long-range dynamics demands an ever-growing number of high-dimensional Gaussians, as shown in Fig.~\ref{fig:figure_page1} top-right. Although all frames are accessible and allow the model to account for disoccluded regions, i.e., areas previously hidden but newly revealed over time, the reconstruction quality in such regions remains limited by restricted representational capacity~\cite{shaw2024swings, kwak2025modec}. Furthermore, practical streaming scenarios require random temporal access to only the relevant portion of the video~\cite{bross2021overview}, yet all-at-once methods still mandate full model transmission, limiting scalability. 

The other candidate is the chunk-based approach (Fig.~\ref{fig:figure_page1}(b), Fig.~\ref{fig:comparison_table}(b)), which partitions long videos into short temporal segments and trains an independent model for each chunk~\cite{wang2024v, li2025gifstream4dgaussianbasedimmersive, liu2025e4dgs}. This divide-and-conquer scheme reduces memory overhead and naturally supports temporal random access. However, optimizing chunks in isolation disrupts temporal consistency, producing boundary artifacts and abrupt appearance shifts when segments are stitched together~\cite{shaw2024swings, li2025gifstream4dgaussianbasedimmersive, wu2025localdygs}. In addition, because each chunk model observes only a limited temporal window, disoccluded regions that emerge in later chunks cannot be reconstructed, reducing overall scene completeness. 

Recent representations tailored for long-range videos still exhibit the trade-offs highlighted in Fig.~\ref{fig:comparison_table}.
The sliding-window strategy (Fig.~\ref{fig:comparison_table}(c))~\cite{shaw2024swings} inherits the benefits of chunk-based methods by training an independent 4DGS model per window, but its overlapping-frame refinement is only a local fix that cannot ensure global temporal consistency, and its reliance on external optical flow increases system complexity.
Another strategy, Temporal Gaussian Hierarchy (Fig.~\ref{fig:comparison_table}(d))~\cite{xu2024longvolcap}, builds a multi-level structure to capture motions at different temporal scales, enabling nearly constant memory usage for long videos. However, maintaining this hierarchy requires continual Gaussian reallocation, per-timestamp segment selection, and CPU–GPU streaming, causing substantially higher system complexity. 

To address the diverse challenges of modeling long-range dynamic content, we propose MoRel, a novel 4DGS framework that overcomes the key limitations of existing methods. First, MoRel introduces the Anchor Relay–based Bidirectional Blending (ARBB) mechanism, which learns bidirectional deformations between Key-frame Anchors (KfA) and blends them through learnable temporal opacity control, effectively suppressing temporal discontinuities. Second, the Feature-variance-guided Hierarchical Densification (FHD) refines anchor representations based on local frequency characteristics, preventing redundant anchor-point generation while preserving high-frequency detail. Third, MoRel maintains bounded GPU memory by dividing long sequences into anchor-based chunking with on-demand loading of only the necessary KfAs and deformation fields. Fourth, the periodic placement of KfA provides natural temporal access points, enabling efficient random temporal access without loading the entire model. Finally, MoRel requires no external cues during training and uses a simple rendering pipeline, avoiding unnecessary system complexity. Together, these components allow MoRel to achieve memory-efficient, flicker-free, and temporally consistent long-range 4D reconstruction suitable for real-world dynamic scenes. In line with these design benefits, quantitative results further show that MoRel outperforms recent 4DGS methods, obtaining the lowest tOF~\cite{chu2020learning}, strong reconstruction quality, and competitive GPU memory usage. 
\vspace{0.1cm}

\vspace{-2mm}
\section{Related Works}
\label{sec:RelatedWorks}


\subsection{All-at-once Approach for 4D NVS}
The first category, the all-at-once approach, jointly optimizes a single canonical representation induced by Gaussian points across all sequence frames, modeling long-range dynamic scenes as a unified space. In 4D Gaussian-based methods~\cite{yang2023deformable3dgs, wu20244d, duan20244d, yang2023gs4d} that explicitly introduce a time index as an extra dimension, the temporal complexity inevitably scales with both the number of Gaussians and the sequence length. As the temporal span increases, these methods often suffer from memory overflow or excessively long training times. Additionally, deformation-based methods~\cite{huang2023sc, deformgs, wang2025freetimegs, li2023spacetime, wu2025swift4d, wu2025localdygs} maintain a canonical representation and learn deformation fields relative to it, achieving significant memory efficiency while retaining reconstruction quality. 
However, such capability is largely a by-product of compactness rather than an intentional design for long-range modeling, and thus still encounters difficulties in handling complex long-range motion video. Moreover, they struggle to address specific challenges, such as newly appearing objects and rapid motions within short-temporal intervals~\cite{song2025coda, lee2024ex4dgs, javed2024tc3d}.

\subsection{Chunk-based Approach for 4D NVS}

Recent Gaussian-based frameworks~\cite{wang2024v, li2025gifstream4dgaussianbasedimmersive, liu2025e4dgs} adopt chunk-based or streamable learning strategies with improved efficiency. For example, GIFStream~\cite{li2025gifstream4dgaussianbasedimmersive} first learns the canonical space and subsequently incorporates the deformation field before introducing compression modules for efficient 4D Gaussian transmission. While these works confirm the scalability potential of Gaussian frameworks, temporal flickering and appearance discontinuities often emerge at chunk boundaries because each segment is optimized independently without explicit inter-chunk consistency modeling.


\begin{figure*}[!t]
\centering
\includegraphics[scale=0.62]{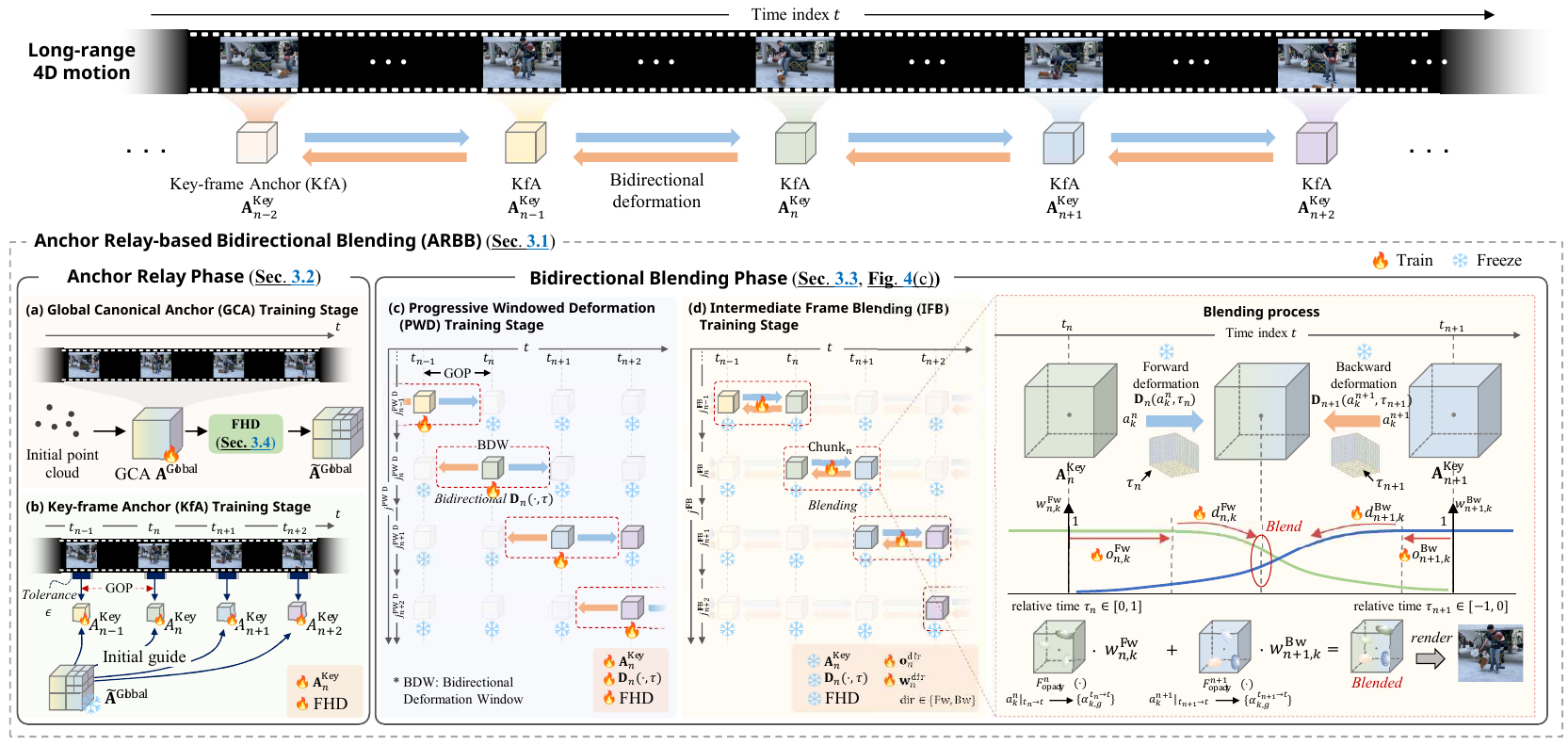}
\vspace{-4mm}
\caption{\textbf{Overview of MoRel framework.} To efficiently model long-range 4D motion with bounded memory and temporal consistency, MoRel adopts the Anchor Relay-based Bidirectional Blending (ARBB) strategy composed of four training stages which are organized into two phases. 
In the Anchor Relay phase (Sec.~\ref{sec:anchor relay}), a GCA is first trained on entire frames with a single point cloud. Next, each KfA is derived around its key-frame time index, while its spatial detail is enhanced through FHD (Sec.~\ref{sec:hierarchical densification}). 
In the Bidirectional Blending phase (Sec.~\ref{sec:bidirectional deformation}), PWD training stage is executed to learn bidirectional deformation fields within local temporal windows to ensure robust motion modeling of each anchor. 
Finally, in IFB training stage, each pair of neighboring anchors is fused through a learnable temporal opacity control, that smoothly transitions anchor influence over time, eliminating temporal flickering across chunks.}
\label{fig:main_figure}
\vspace{-3mm}
\end{figure*}

\section{Proposed Method}
\label{Method}

\subsection{Overview of MoRel}
\label{sec:overview of morel}
We adopt the anchor-point–based representation \cite{lu2024scaffold}, where a sparse voxel grid of anchor-points defines a canonical space parameterized by neural Gaussians. The preliminaries and \textbf{all notations used throughout the paper are provided in \textit{Suppl.~\ref{sec:notation}}}. 
Building on this representation, our goal is to model long-range 4D motion with bounded memory usage while maintaining temporal coherence and motion fidelity, toward real-world system applicability. 
To this end, we employ the Anchor Relay-based Bidirectional Blending (ARBB) strategy (Sec.~\ref{sec:Introduction}, Fig.~\ref{fig:figure_page1}), which consists of two sequential phases illustrated in Fig.~\ref{fig:main_figure} and is further organized into four training stages. This design enables scalable and efficient long-range 4D motion modeling under bounded memory conditions.
In the Anchor Relay phase (Sec.~\ref{sec:anchor relay}), training begins with (i) the Global Canonical Anchor (GCA) training stage (Fig.~\ref{fig:main_figure}(a)), where a single GCA is trained from entire frames to provide a globally consistent initialization. 
After training the GCA, variance-based levels are respectively assigned to anchor-points for later use in the Feature-variance-guided Hierarchical Densification (FHD) (Sec.~\ref{sec:hierarchical densification}).
Subsequently, (ii) the Key-frame Anchor (KfA) training stage (Fig.~\ref{fig:main_figure}(b)) optimizes each temporally periodic KfA at a finer level respectively induced from the level-assigned GCA, forming local canonical spaces optimized for their respective chunks.
Each KfA is further refined by FHD with the pre-assigned level to enhance spatial detail while suppressing redundant Gaussians, efficiently balancing memory overhead and reconstruction quality. 
In the Bidirectional Blending phase (Sec.~\ref{sec:bidirectional deformation}), bidirectional deformation and temporal blending are learned through two consecutive stages. 
(i) In the Progressive Windowed Deformation (PWD) training stage (Fig.~\ref{fig:main_figure}(c), each key anchor \textit{independently} learns forward and backward deformation fields within sliding temporal windows, \textit{bounding memory usage} and preventing inter-chunk interference. 
Next, (ii) the Intermediate Frame Blending (IFB) stage (Fig.~\ref{fig:main_figure}-(d)) learns a temporal fusion model through a learned temporal opacity control, which adaptively transits KfA influence over time to achieve smooth and flicker-free motion continuity. 
By integrating these two phases, MoRel achieves high-fidelity, temporally coherent, and memory-efficient long-range 4D motion reconstruction.

\subsection{Anchor Relay Phase}
\label{sec:anchor relay}

\noindent \textbf{(i) Global Canonical Anchor Training.}\quad 
In this stage, we skim through the entire video sequence to train $\textbf{A}^{Global}$ as shown in Fig.~\ref{fig:main_figure}(a). 
It serves to ensure global consistency across the entire sequences.
In addition, we observed that previous works \cite{sun20243dgstream, wang2025freetimegs,li2025gifstream4dgaussianbasedimmersive} require temporally-dense point clouds for modeling 4D motion, such as using all frames or sampling one every ten frames. 
However, for a long-range 4D motion, such frequent point cloud requirements can significantly increase computational and memory overhead. 
To address this, we construct a single point cloud for the initialization points for $\textbf{A}^{\text{Global}}$ detailed in  \textit{Suppl.~\ref{initial point cloud}}. Such an efficient initialization not only ensures global coherence but also makes the framework be practically applicable to long-range 4D scenes.
After training $\textbf{A}^{\text{Global}}$, feature variance-based levels $L_{a^{\text{Global}}_k}$ are assigned to its each anchor-point ${a^{\text{Global}}_k}$ by Eq.~\ref{eq:level assignment}, resulting  $\mathbf{\widetilde{A}}^{\text{Global}}$.


\noindent \textbf{(ii) Key-frame Anchor Training.}\quad
To effectively handle long-range 4D motion, we newly introduce periodically placed KfA, which are finely optimized around their corresponding time indices $t_n$, enabling high-quality reconstruction of the assigned key moment. 
All $\mathbf{A}_n^{\text{Key}}$ are initialized from the level-assigned global anchor $\mathbf{\widetilde{A}}^{\text{Global}}$, rather than being trained from scratch, which ensures globally consistent appearance across the sequence.
Each $\mathbf{A}_n^{\text{Key}}$ also serves as a local canonical space within its associated temporal range, $[\,\max(0,\, t_n - \mathrm{GOP}),\, \min(t_n + \mathrm{GOP},\, T-1)\,]$, where GOP (Group-of-Pictures) denotes the temporal spacing between adjacent KfAs.
To enhance robustness, we introduce a temporal tolerance $\epsilon$, allowing each $\mathbf{A}_n^{\text{Key}}$ to capture variations within its local temporal $\epsilon$-neighborhood, $[\,\max(0,\, t_n - \epsilon),\, \min(t_n + \epsilon,\, T-1)\,]$.
In addition, to further refine spatial detail while suppressing redundant Gaussians, we employ FHD (Sec.~\ref{sec:hierarchical densification}), which adaptively grows and prunes anchor-points based on the frequency characteristics of each $a_k^n\in \textbf{A}^{\text{Key}}_n$ using the pre-assigned  $L_{a^{\text{Global}}_k}$.
The periodic placement of KfAs, inspired by~\cite{sullivan2012overview, bross2021overview}, improves practical applicability in real-world systems 
by providing random access points and maintaining bounded memory usage, as detailed in  Alg.~\ref{alg:MoRel_training}, \ref{alg:MoRel_rendering}.

\subsection{Bidirectional Blending Phase}
\label{sec:bidirectional deformation}

\begin{figure}[t!]
\centering
\includegraphics[scale=0.53]{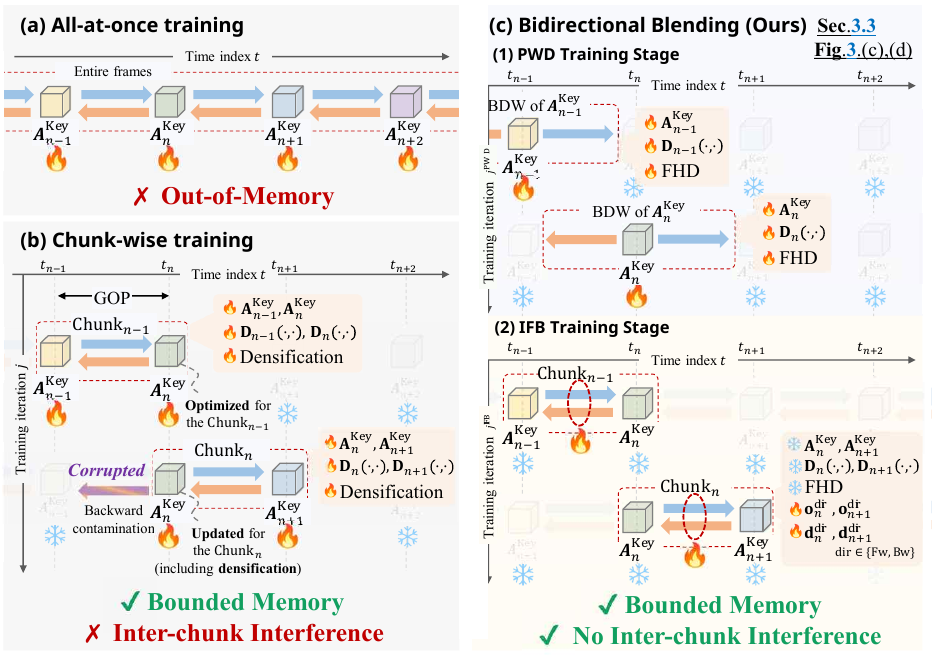}
\caption{\textbf{Comparison of training strategies for modeling long-range 4D motion with bidirectional deformation}. (a) All-at-once training suffers from memory overflow. (b) Chunk-wise training reduces memory cost but causes inter-chunk interference. (c) Our Bidirectional Blending (PWD + IFB) maintains bounded memory and prevents inter-chunk interference.}
\label{fig:progressive training}
\vspace{-2mm}
\end{figure}

\noindent \textbf{(i) Progressive Windowed Deformation Training.}\quad As introduced in Fig.~\ref{fig:figure_page1}, we adopt a bidirectional deformation scheme to better handle irregular motion and to achieve smooth transitions across adjacent KfAs. 
However, directly optimizing long-range 4D motion under this bidirectional design remains challenging as shown in Fig.~\ref{fig:progressive training}. 
An all-at-once training (Fig.~\ref{fig:progressive training}(a)) suffers from severe memory explosion. 
Alternatively, chunk-wise training (Fig.~\ref{fig:progressive training}(b)) can alleviate 
this memory issue. Here, a chunk refers to a temporal frame range that can be rendered without reloading a new KfA. However, this scheme can cause an inter-chunk interference problem.
Specifically, in the case of (Fig.~\ref{fig:progressive training}(b), $\mathbf{A}_n^{\text{Key}}$ is trained for the $\text{chunk}_{n-1}$, but later updated again for the $\text{chunk}_{n}$ including densification. Accordingly, the characteristics previously optimized for $\text{chunk}_{n-1}$ become corrupted. For example, new anchor-points may grow that were not trained for backward deformation toward $\text{chunk}_{n-1}$, or existing anchor-points crucial for representing $\text{chunk}_{n-1}$ may be pruned during the training for $\text{chunk}_{n}$. We refer this phenomenon to \textit{backward contamination}. To address this, we propose the PWD training strategy as illustrated in Fig.~\ref{fig:progressive training}(c).
In our PWD training, each KfA is \textit{independently} optimized within a local Bidirectional Deformation Window (BDW), defined as a temporal window modeled by a single KfA via bidirectional deformation. 
Each $\mathbf{A}_n^{\text{Key}}$ is dynamically loaded only when its $\text{BDW}_n$ is being optimized and unloaded afterward described in Alg.~\ref{alg:MoRel_training}, resulting in efficient bounded memory usage, i.e., \textit{on-demand loading}.
Each $\text{BDW}_n$ is trained for $J_n^{\text{{PWD}}}$ iterations and the window then progressively slides to the next BDW with overlapping a chunk.
This progressive training not only bounds memory usage but also prevents inter-chunk interference, enabling stable and consistent long-range motion learning. 
Within the BDW, the deformation field $\textbf{D}_n(\cdot,\tau_n)$ of $\textbf{A}_n^{\text{Key}}$ performs both forward $(+)$ and backward $(-)$ deformation using the normalized relative time $\tau_n \in[-1,1]$ corresponding to $t\in[t_n-\text{GOP}, t_n+\text{GOP}]$. Each $a_k^n$ queries its position to $\textbf{D}_n(\cdot,\tau_n)$ to obtain the deformation amount of its attribute, further detailed in \textit{Suppl.~\ref{Preliminary:Hexplane}}.

\noindent \textbf{(ii) Intermediate Frame Blending Training.}\quad 
After the deformation fields are established through PWD, we proceed to the IFB training stage, where two adjacent KfAs, $\textbf{A}_n^{\text{Key}}$ and $\textbf{A}_{n+1}^{\text{Key}}$, are \textit{jointly loaded} to train blending mechanism to smoothly blend the resulting KfAs for $J_n^{IFB}$ iterations within the corresponding $\text{chunk}_n$. 
In here, as shown in Fig.~\ref{fig:main_figure}(d), Fig.~\ref{fig:progressive training}(c), only the blending weights are trained while the anchor attributes, deformation fields, and FHD remain frozen.
For the blending weight, we build upon the opacity control mechanism adopted in previous works \cite{li2024spacetime, duan20244d, wang2025freetimegs}, where opacity exponentially decays with the temporal distance from a central time which is $t_n$ for $\textbf{A}_n^{\text{Key}}$ in our case.
However, in dynamic scenes with irregular motion such as occlusion, the spatio-temporal influence of each KfA can vary non-uniformly. 
To effectively model these characteristics, we newly introduce a learnable temporal opacity control, assigning each anchor-point $a_k^n$ its own temporal offset $o_{n,k}^{\text{dir}}$, and temporal decay speed $d_{n,k}^{\text{dir}}$, where $\text{dir}\in\{\text{Fw, Bw}\}$ (forward and backward).
As shown in Fig.~\ref{fig:main_figure}(d), the temporal opacity of $n$-th KfA is defined as 
\begin{equation} \label{eq:opacity control}
    w_{n,k}^{\text{dir}} = \text{exp}[-\lambda_{\text{decay}}\cdot d_{n,k}^{\text{dir}}\cdot|\tau_n-o_{n,k}^{\text{dir}}|],
\end{equation}
where $\lambda_{\text{decay}}$ denotes a base decay coefficient that controls the global decay speed. This learnable opacity control-based bidirectional blending enables not only smooth transitions between adjacent KfAs but also robust representation of irregular motions in the dynamic scene. 
The effectiveness of this design is validated through ablation studies (Tab.~\ref{table:ablation_study}). 
Using the learned weight, the blending process is performed after reconstructing neural Gaussians \cite{lu2024scaffold} of each deformed anchor $a^n_k|_{t_n\to t}$. The obtained opacity values at $a^n_k|_{t_n\to t}$ and $a^{n+1}_k|_{t_{n+1}\to t}$ are blended according to the learned weights rendered \cite{kerbl20233d} to produce the final output. 
The overall processing pipeline, from anchor-points to the rendered image including the blending process, is visualized and illustrated in \textit{Suppl.~\ref{fig:Overview of processing pipeline}}.

\subsection{Feature-variance-guided Hierarchical Densification}
\label{sec:hierarchical densification}
\begin{figure}[t!]
\centering
\includegraphics[scale=0.37]{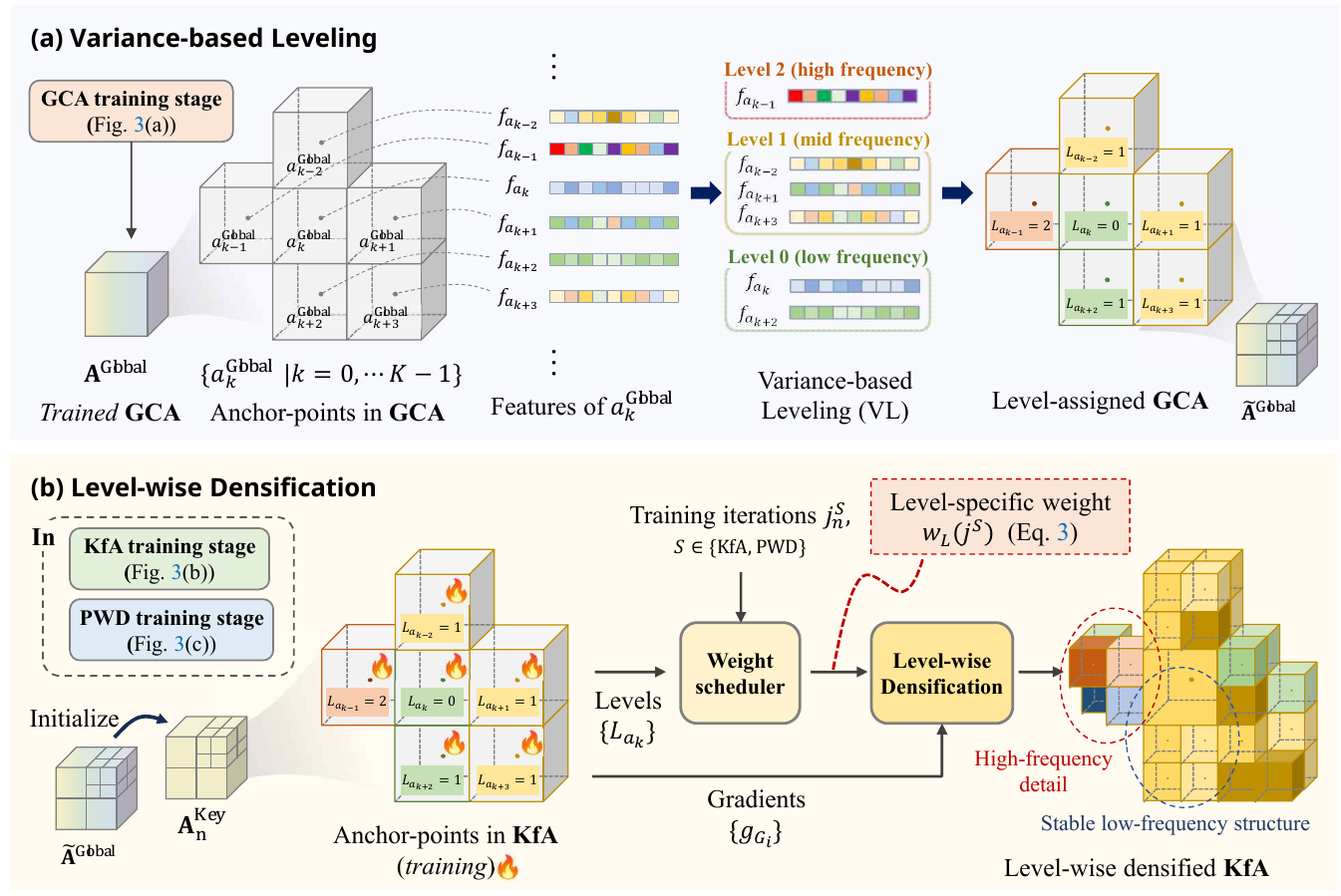}
\vspace{-1mm}
\caption{\textbf{Overview of Feature-variance-guided Hierarchical Densification.} (a) Variance-based Leveling: After GCA training, we assign a level to each anchor-point guided by the feature-variance. (b) Level-wise Densification: During the KfA and PWD trainings, gradients for KfA densification are modulated by level-specific weights, enabling early low-frequency stabilization and late high-frequency refinement.}
\label{fig:FHD}
\end{figure}

In FHD, we use anchor-point's feature $\hat{f}_k$ variance as a proxy of local frequency complexity and modulate level-wise densification. As shown in Fig.~\ref{fig:FHD}, the module comprises Variance-based Leveling (VL) and Level-wise Densification (LD). The objective is to suppress redundant anchor growth in unstable high-frequency regions during early training iterations and to enhance fine detail in later iterations of the KfA and PWD Training stages. 

\noindent\textbf{Variance-based Leveling.}
After completing GCA training, each global anchor-point $a^{\text{Global}}_k$ in $\textbf{A}^{\text{Global}}$ is assigned a hierarchical level that reflects local frequency complexity, resulting in the level-assigned global anchor set $\mathbf{\widetilde{A}}^{\text{Global}}$.
The feature representation $\hat{f}_k$ of each global anchor-point exhibits varying sensitivity depending on local frequency characteristics. 
High-frequency components are better fitted during later training stages and are particularly sensitive in the early phase~\cite{rahaman2019spectral}. 
Repeated accumulation of large gradient fluctuations increases the variance of $\hat{f}_k$~\cite{qian2020impact}. 
Therefore, the variance of anchor features $\sigma_k^2 = \mathrm{Var}(\hat{f}_k)$ serves as a reliable indicator of local frequency complexity. 
For each anchor-point $a^{\text{Global}}_k$, hierarchical levels $L_{a^{\text{Global}}_k}$ are assigned based on quantile thresholds $\{\tau_1, \tau_2\}$ as follows:
\begin{equation}
L_{a^{\text{Global}}_k} =
\begin{cases}
0, & \sigma_k^2 < \tau_1 \quad (\text{low-frequency}),\\[2pt]
1, & \tau_1 \leq \sigma_k^2 < \tau_2,\\[2pt]
2, & \sigma_k^2 \geq \tau_2 \quad (\text{high-frequency}).
\end{cases}
\label{eq:level assignment}
\end{equation}
The $L_{a^{\text{Global}}_k}$ is then used as a control signal for LD, enabling balanced densification in terms of memory and high-frequency detail. Ablation and analysis on the number of levels along with visualization between anchor-point feature and image frequency are provided in \textit{Suppl.~\ref{Analysis on FHD}}. 

\noindent\textbf{Level-wise Densification.}
LD makes growth decisions based on gradient statistics at each level. 
Following prior gradient-based densification schemes in anchor-based representations~\cite{lu2024scaffold}, we track the accumulated gradient for each neural gaussian instead of using a single-step gradient. 
At training iteration $j^\mathcal{S}_n$, where $S \in \{\text{KfA}, \text{PWD}\}$ denotes the current training stage, $g^{j^\mathcal{S}_n}$ represents the magnitude of the gradient accumulated up to iteration $j^\mathcal{S}_n$.
We apply a level-specific weight $w_L^{j^{\mathcal{S}}_n}$, forming the level-weighted statistic $g_L^{j^{\mathcal{S}}_n} = g^{j^{\mathcal{S}}_n} \cdot w_L^{j^{\mathcal{S}}_n}$, which serves as the criterion for level-wise densification.
Here, $w_L^{j^\mathcal{S}_n}$ adjusts the relative importance of level $L$ at training iteration $j^\mathcal{S}_n$, placing greater emphasis on lower levels during early training to stabilize low-frequency structures, and gradually increasing the weight of higher levels in later stages to enhance high-frequency details.
Specifically, $w_L^{j^\mathcal{S}_n}$ follows a linear interpolation between the initial and final importance factors as
\begin{equation}
w_L^{j^\mathcal{S}_n} =
\begin{cases}
1, & L = 0, \\
\lambda_L + (1 - \lambda_L)\eta_t, & L \ge 1,
\end{cases}
\label{eq:densification}
\end{equation}
where $\eta_t = \frac{j^\mathcal{S}_n}{J^\mathcal{S}_n}$ denotes the normalized training progress ratio, and $J^\mathcal{S}_n$ represents the total number of training iterations for each $\mathbf{A}^{\text{Key}}_n$ in the corresponding stage. 
Here, $\lambda_L$ controls the initial importance assigned to level $L$, enabling lower levels to retain stronger weights at early iterations while allowing higher levels to gain influence as $\eta_t$ increases.
This formulation ensures a smooth temporal transition from low-frequency–dominated densification at the beginning to high-frequency–oriented refinement toward the end. Neural gaussians that satisfy the growth criterion based on $g_L^{j^\mathcal{S}_n}$ are mapped onto the spatial grid and used as candidate positions for new anchors.

\vspace{-1mm}
\subsection{Summary of Training and Rendering Process}
\label{sec:summary}
\vspace{-1mm}
The overall training procedure of our \textbf{MoRel} is summarized in Alg.~\ref{alg:MoRel_training}.
As described in Sec.~\ref{sec:overview of morel}, Fig.~\ref{fig:main_figure}, four training stages are sequentially performed, and each training stage is iteratively optimized for a predefined stage-specific number of iterations $J^S,S\in\{\text{GCA,KfA,PWD,IFB}\}$.
Note that, through the \textit{dynamic loading} and \textit{unloading} operations indicated in the algorithm, only one or two key anchors and their corresponding deformation fields are loaded at any given time, \textit{ensuring bounded memory usage during training}.
Similarly, the rendering process summarized in Alg.~\ref{alg:MoRel_rendering} utilizes the same bounded-memory principle. 
For a given rendering time $t$, the required KfAs are first determined based on the GOP, and loaded only when necessary for rendering. 
This on-demand loading minimizes redundant memory usage, minimizing unnecessary load/unload operations. 
Through this design, \textbf{MoRel} achieves memory-efficient and temporally consistent spatio-temporal novel view synthesis for long-range 4D motion.

\begin{algorithm}[t]
\small
\caption{Training process of MoRel (Fig.~\ref{fig:main_figure})}
\begin{algorithmic}[1]
\Procedure{Train MoRel}{}
\State \textbf{Input} point cloud $\mathcal{P}$, cameras $\mathbf{C}=\{\, c^m_t \mid \substack{m=0,\dots,M-1 \\ t=0,\dots,T-1} \,\}$, GOP, $J^\mathcal{S}$ where $\mathcal{S}\in \{\text{GCA}, \text{KfA}, \text{PWD}, \text{IFB}\}$, tolerance $tol$
    \State \process{\textbf{GCA Training Stage}} (Sec.~\ref{sec:anchor relay}(i), Fig.~\ref{fig:main_figure}(a))
    \State Initialize $\mathbf{A}^{\text{Global}}$ by $\mathcal{P}$
    \For{$j^{\text{GCA}}$ in $J^{\text{GCA}}$}
        \State Pick random $c_t^m \in \textbf{C}$
        \State \textbf{Train} $\mathbf{A}^{\text{Global}}$ with $c_t^m$
    \EndFor
    \State $L_{a^\text{Global}}$ is pre-assigned to $\mathbf{A}^{\text{Global}}$ for \textbf{FHD} (Sec.~\ref{sec:hierarchical densification})
    \State \process{\textbf{KfA Training Stage}} (Sec.~\ref{sec:anchor relay}(ii), Fig.~\ref{fig:main_figure}(b))
    \State Initialize $\{\mathbf{A}_n^{\text{Key}}\}_{n \in [0,N-1]}$ from $\mathbf{A}^{\text{Global}}$,  $N{=}\lceil T/\mathrm{GOP}\rceil$
    \For{$n \in [0,N-1]$}   
        \State \load{\textbf{Load}} $\mathbf{A}_n^{\text{Key}}$
        \For{$j_n^{\text{KfA}} \in J_n^{\text{KfA}} $}
            \State Pick random $\{c^m_t \in \textbf{C}\mid \substack{t \in [\,t_n - tol,\, t_n + tol\,] \\t \in [\,0,\, T-1\,]} \,\}$
            \State \textbf{Train} $\mathbf{A}_n^{\text{Key}}$ with $c_t^m$, \textbf{densified} by \textbf{FHD} 
        \EndFor
        \State \unload{\textbf{Unload}} $\mathbf{A}_n^{\text{Key}}$
    \EndFor

    \State \process{\textbf{PWD Training Stage}} (Sec.~\ref{sec:bidirectional deformation}(i), Fig.~\ref{fig:main_figure}(c))
    \For{$n \in [0,N-1]$}   
        \State \load{\textbf{Load}} $\mathbf{A}_n^{\text{Key}}, \textbf{D}_n(\cdot,\tau_n)$
        \For{$j_n^{\text{PWD}} \in J_n^{\text{PWD}} $}
        \State Pick random $\{c^m_t \in \mathbf{C}\mid
        \smash{\overset{\text{\scriptsize \frozen{BDW}}_n}{
        \substack{
        t \in [\,t_n - \text{GOP},\, t_n + \text{GOP}\,] \\
        t \in [\,0,\, T-1\,]
        }}}
        \,\}$
            \State \textbf{Train} $\mathbf{A}_n^{\text{Key}}, \textbf{D}_n(\cdot,\tau_n)$ with $c_t^m$, \textbf{densified} by \textbf{FHD} 
        \EndFor
         \State \unload{\textbf{Unload}} $\mathbf{A}_n^{\text{Key}}, \textbf{D}_n(\cdot,\tau_n)$
    \EndFor

    \State \process{\textbf{IFB Training Stage}} (Sec.~\ref{sec:bidirectional deformation}(ii), Fig.~\ref{fig:main_figure}(d))
    \For{$n \in [0,N-2]$}   
        \State \load{\textbf{Load}} $\mathbf{A}_n^{\text{Key}}, \mathbf{A}_{n+1}^{\text{Key}}, \textbf{D}_n(\cdot,\tau_n), \textbf{D}_{n+1}(\cdot,\tau_{n+1})$
        \For{$j_n^{\text{IFB}} \in J_n^{\text{IFB}} $}
            \State Pick random $\{c^m_t \in \mathbf{C}\mid
            \smash{\overset{\text{\scriptsize \frozen{Chunk}}_n}{
            \substack{
            t \in [\,t_n,\, t_n + \text{GOP}\,] \\
            t \in [\,0,\, T-1\,]
            }}}
            \,\}$
            \State \textbf{Train} $o_n^\text{Fw}, w_n^{\text{Fw}},o_{n+1}^\text{Bw}, w_{n+1}^{\text{Bw}}$,
            \State with frozen $\mathbf{A}_n^{\text{Key}}, \mathbf{A}_{n+1}^{\text{Key}}, \textbf{D}_n(\cdot,\tau_n), \textbf{D}_{n+1}(\cdot,\tau_{n+1})$ 
        \EndFor
        \State \textbf{Save} $\mathbf{A}_n^{\text{Key}}, \mathbf{A}_{n+1}^{\text{Key}}, \textbf{D}_n(\cdot,\tau_n), \textbf{D}_{n+1}(\cdot,\tau_{n+1}), o_n^\text{dir}, w_n^{\text{dir}}$
        \State \unload{\textbf{Unload}} $\mathbf{A}_n^{\text{Key}}, \mathbf{A}_{n+1}^{\text{Key}}, \textbf{D}_n(\cdot,\tau_n), \textbf{D}_{n+1}$
    \EndFor

\EndProcedure
\end{algorithmic}
\label{alg:MoRel_training}
\end{algorithm}


\begin{algorithm}[h]
\small
\caption{Rendering process of MoRel}
\begin{algorithmic}[1]
\Procedure{Render MoRel}{$c^{\text{ren}}_t$, GOP}
    \State $n=\lfloor t/\text{GOP} \rfloor$
    \vspace{1mm}
    \If{$\mathbf{A}_n^{\text{Key}}, \mathbf{A}_{n+1}^{\text{Key}}, \textbf{D}_n(\cdot,\tau_n), \textbf{D}_{n+1}$ are \train{\textbf{not}} loaded} 
        \State \load{\textbf{Load}} {$\mathbf{A}_n^{\text{Key}}, \mathbf{A}_{n+1}^{\text{Key}}, \textbf{D}_n(\cdot,\tau_n), \textbf{D}_{n+1}$}
    \EndIf
    \State \textbf{Render} $c^{\text{ren}}_t$ by Bi-directional Blending \\
    \quad\quad with $\mathbf{A}_n^{\text{Key}}, \mathbf{A}_{n+1}^{\text{Key}}, \textbf{D}_n(\cdot,\tau_n), \textbf{D}_{n+1}, o_n^\text{dir}, w_n^{\text{dir}}$
        \vspace{1mm}
    \If{$t = (n+1)\cdot\text{GOP}$}
        \State \unload{\textbf{Unload}} $\mathbf{A}_n^{\text{Key}}, \mathbf{A}_{n+1}^{\text{Key}}, \textbf{D}_n(\cdot,\tau_n), \textbf{D}_{n+1}, o_n^\text{dir}, w_n^{\text{dir}}$
    \EndIf
\EndProcedure
\end{algorithmic}
\label{alg:MoRel_rendering}
\end{algorithm}

\vspace{-1mm}
\section{Experiments}
\label{Experiments}

\subsection{Experimental setup}
\label{experimental setup}
\noindent\textbf{Implementation Details.}
\label{implementation detail}
MoRel is built upon 4DGS~\cite{wu20244d} and Scaffold-GS~\cite{lu2024scaffold}, retaining most hyperparameters. Implementation details are provided in \textit{Suppl.~\ref{Preliminary}}.


\noindent\textbf{$\textbf{SelfCap}_{LR}$}.
\label{Datasets}To simulate real-world long-range 4D motion, we construct $\textbf{SelfCap}_{LR}$ dataset selected from an undistilled raw video dataset~\cite{wang2025freetimegs,xu2024longvolcap}. It contains five representative and challenging dynamic sequences (Bike$_{1}$, Bike$_{2}$, Corgi, Yoga, and Dance) over 3500 frames. It has larger average motion magnitude and captured at wider spaces compared to other previous datasets as shown in \textit{Suppl.~\ref{Dataset Statistics}}. As a result, it is an appropriate benchmark to evaluate challenging long-range 4D motion modeling.

\noindent\textbf{Evaluation Metrics.} We employ PSNR, SSIM, and LPIPS~\cite{zhang2018unreasonable}. To further evaluate temporal consistency, we adopt tOF~\cite{chu2020learning} between consecutive frames, as detailed in \textit{Suppl.~\ref{tOF}}. We additionally measure training and rendering memory, which are crucial evaluation factors for long-range 4D motion modeling in terms of efficiency.


\subsection{Comparisons}
\noindent \textbf{Comparison methods.} To validate our model, we conducted comparisons against state-of-the-art (SOTA) methods, which can be categorized into the (i) all-at-once training and (ii) chunk-based training introduced in Sec.~\ref{sec:Introduction}, \ref{sec:RelatedWorks}.
For the all-at-once training category, we included 4DGS \cite{wu20244d}, MoDec-GS~\cite{kwak2025modec}, and LocalDyGS \cite{wu2025localdygs}, while GIFStream \cite{li2025gifstream4dgaussianbasedimmersive} was used as a representative chunk-based method.
In addition, for a more comprehensive evaluation, we adapted 4DGS to operate in a chunk-wise manner and included it in our comparison experiments, named 4DGS$_{\text{chunk}}$. Also, demo videos and additional benchmark results on the DyCheck-iPhone, HyperNeRF, DyNeRF, and PanopticSports datasets are provided in \textit{Suppl.~\ref{Additional Results}}.

\begin{figure}[!h]
\centering
\vspace{-3mm}
\includegraphics[scale=0.47]{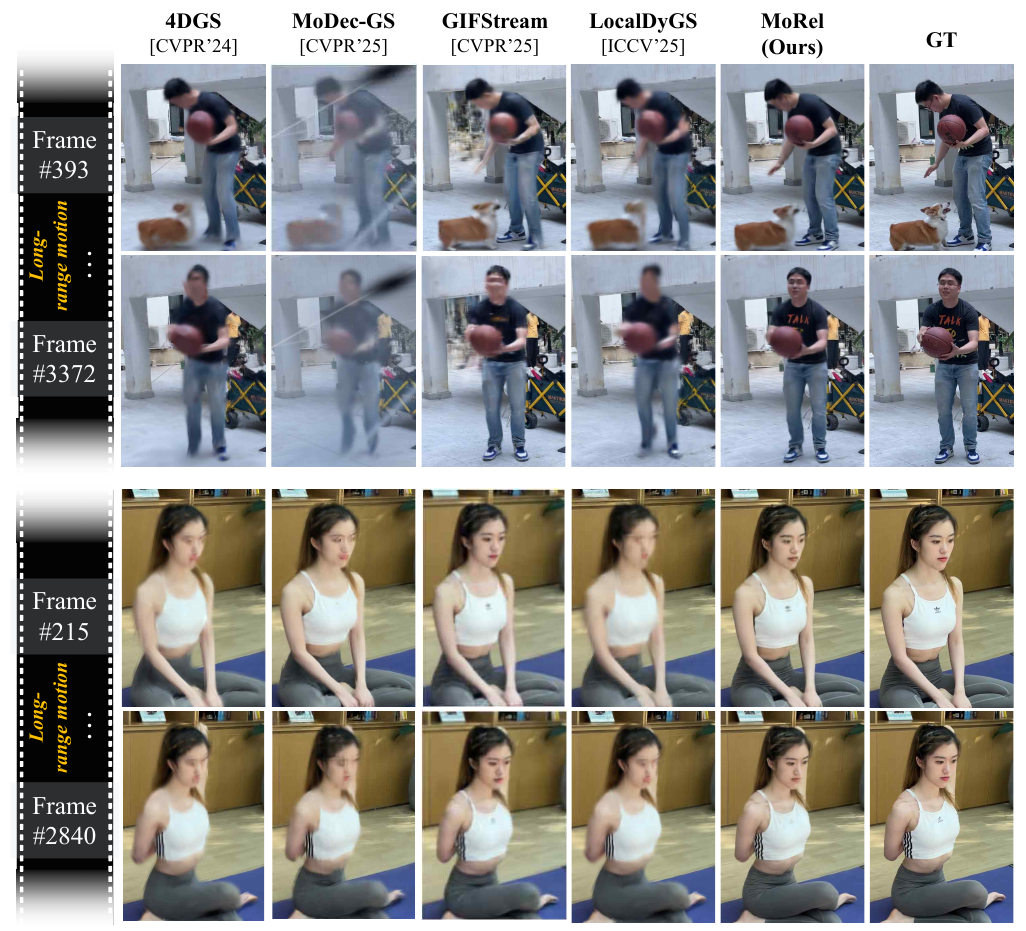}
\vspace{-3mm}
\caption{\textbf{Qualitative comparison on SelfCap$_\text{LR}$.} Our \textbf{MoRel} demonstrates superior visual fidelity in long-range motion modeling compared to existing SOTA methods, thanks to its ARBB mechanism that effectively handles long-range 4D motion.}
\label{fig:qualitative results}
\vspace{-5mm}
\end{figure}

(i) \textbf{All-at-once training.} As shown in Tab.~\ref{table:selfcap_comparison}, all-at-once approaches exhibit generally lower quantitative results than ours, indicating degraded structural accuracy and visual fidelity. 
Our method shows particularly strong advantages on scenes with spatio-temporally large motion such as Corgi, Yoga and Dance, consistently outperforming the existing methods. 
These trends are further verified in the qualitative comparison shown in Fig.~\ref{fig:qualitative results}. For long-range 4D motion with 2k or more frame range, all-at-once approaches tend to lose motion expressiveness and degrade fine details due to their brute-force global optimization. In addition, these methods also suffer from the memory explosion problem, as shown in Tab.~\ref{table:selfcap_tof_memory}, which interferes the practical applicability to real-world systems. 
In contrast, our MoRel delivers fine and robust reconstruction under even long-range motion with bounded memory usage thanks to the ARBB mechanism (Sec.~\ref{sec:overview of morel}, Fig.~\ref{fig:main_figure}).

(ii) \textbf{Chunk-based training.}
Chunk-based approaches~\cite{li2025gifstream4dgaussianbasedimmersive} similarly exhibit overall lower and relatively unstable quantitative performance than our method, as shown in Tab.~\ref{table:selfcap_comparison}. 
In particular, our adapted 4DGS shows reasonable performance on relatively static scenes such as Bike$_{1}$ and Bike$_{2}$, with a tendency to maintain SSIM. However, as shown in Tab.~\ref{table:selfcap_tof_memory}, it suffers from substantially worse tOF scores. This indicates degraded temporal consistency caused by flickering at chunk boundaries. This observation is further supported by the temporal profile analysis provided in \textit{Suppl.~\ref{Temporal Profile Analysis}}. In contrast, our method benefits from bidirectional blending (Sec.~\ref{sec:bidirectional deformation}), achieving not only superior quantitative results in Tab.~\ref{table:selfcap_comparison} but also the best tOF score in Tab.~\ref{table:selfcap_tof_memory} among all compared approaches.

\begin{table*}[!h]
\begin{center}
\setlength\tabcolsep{3pt} 
\renewcommand{\arraystretch}{1.25} 
\scalebox{0.7}{ 
\begin{tabular}{ c | c | c | c | c  | c  | c || c}
\bottomrule \hline
\noalign{\smallskip}
Group & Method & Bike$_1$ & Bike$_2$ & Corgi & Yoga & Dance & \textbf{Average} \\  
\hline\noalign{\smallskip}

\multirow{3}{*}{(a)} 
  & 4DGS [CVPR'24]~\cite{wu20244d}   
  & 21.62 / 0.662 / 0.375 
  & \second{22.61} / \best{0.700} / 0.346 
  & 19.72 / 0.566 / 0.489 
  & 13.72 / 0.715 / 0.402 
  & 17.09 / 0.597 / 0.399 
  & 18.95 / 0.648 / 0.402  \\ 

  & MoDec-GS [CVPR'25]~\cite{kwak2025modec}  
  & 21.55 / 0.659 / 0.355  
  & 22.13 / 0.666 / 0.348  
  & 17.38 / 0.539 / 0.492  
  & 17.51 / 0.732 / 0.396  
  & 17.48 / \second{0.621} / \best{0.363}   
  & 19.61 / 0.643 / 0.391 \\ 

  & LocalDyGS [ICCV'25]~\cite{wu2025localdygs}  
  & 22.25 / 0.664 / \second{0.341} 
  & \best{22.66} / 0.675 / \second{0.340}  
  & 19.37 / 0.551 / \second{0.462}  
  & 21.38 / 0.755 / 0.333  
  & \second{17.56} / 0.615 / \second{0.377}  
  & \second{20.64} / 0.652 / \second{0.371} \\ 

\hline

\multirow{2}{*}{(b)}
  & GIFStream [CVPR'25]~\cite{li2025gifstream4dgaussianbasedimmersive}  
  & 18.43 / 0.658 / 0.345 
  & 20.11 / 0.655  / 0.345  
  & 19.83 / 0.566 / 0.467 
  & \second{22.02} / \second{0.771} / \second{0.335}  
  & 14.73 / 0.614 / 0.533 
  & 19.02 / 0.653 / 0.405 \\ 

  & $\text{4DGS}_\text{chunk}$
  & \second{22.26} / \best{0.695} / 0.344 
  & 22.52 / \second{0.699} /  \second{0.340} 
  & \second{19.90} / \second{0.570} / 0.492 
  & 14.62 / 0.712 / 0.389 
  & 17.23 / 0.603 / 0.379 
  & 19.31 / \second{0.656} / 0.389 \\ 

\hline

(c) & MoRel (\textbf{Ours})
  & \best{22.32} / \second{0.668} / \best{0.321}  
  & 22.57 / 0.670 / \best{0.319}  
  & \best{19.93} / \best{0.575} / \best{0.461}  
  & \best{22.18} / \best{0.780} / \best{0.297}  
  & \best{17.99} / \best{0.628} / \second{0.377}  
  & \best{21.00} / \best{0.664} / \best{0.355} \\ 

\bottomrule \hline
\end{tabular}
}
\vspace{-2mm}
\caption{\textbf{Quantitative results comparison on our newly composed SelfCap$_{\text{LR}}$.} Group denotes (a) all-at-once training methods, (b) chunk-based approaches including our unidirectional deformation variant, and (c) our \textbf{MoRel} model.
\textcolor{red}{\textbf{Red}} and \textcolor{blue}{\underline{blue}} denote the best and second-best performances, respectively. 
Each block element of 3-performance denotes (PSNR (dB)$\uparrow$ / SSIM$\uparrow$ / LPIPS$\downarrow$).} 
\label{table:selfcap_comparison}
\end{center}
\vspace{-4mm}
\end{table*}


\begin{table}[!h]
\begin{center}
\setlength\tabcolsep{4pt} 
\renewcommand{\arraystretch}{1.3} 
\scalebox{0.75}{ 
\begin{tabular}{ c | c | c | c c }
\bottomrule \hline
\noalign{\smallskip}
\multirow{2}{*}{Group} & 
\multirow{2}{*}{Method} & 
\multirow{2}{*}{tOF$\downarrow$} & 
\multicolumn{2}{c}{Memory (MB)$\downarrow$} \\  
\cline{4-5}\noalign{\smallskip}
& & & Training & Rendering \\
\bottomrule
\hline\noalign{\smallskip}

\multirow{3}{*}{(a)} 
  & 4DGS [CVPR'24]~\cite{wu20244d}   & 0.222 & $\sim$18,000 & 143 \\
  & MoDec-GS [CVPR'25]~\cite{kwak2025modec}  & 0.249 & $\sim$22,000 & 154 \\
  & LocalDyGS [ICCV'25]~\cite{wu2025localdygs}  & \second{0.215} & $\sim$12,000 & 122 \\
\hline

\multirow{2}{*}{(b)}
  & GIFStream [CVPR'25]~\cite{li2025gifstream4dgaussianbasedimmersive}  & 0.539 & $\sim$9,000 & 93 \\
  & $\text{4DGS}_\text{chunk}$  & 0.680 & $\sim$4,500 & 65 \\
\hline

(c) & MoRel (\textbf{Ours})  & \best{0.203} & $\sim$6,000 & 126 \\
\bottomrule \hline
\end{tabular}
}
\vspace{-1mm}
\caption{\textbf{Metrics critical to long-range motion modeling.}
We highlight the key factors that determine a model’s capability in long-range motion handling} 
\label{table:selfcap_tof_memory}
\end{center}
\vspace{-1mm}
\end{table}


\vspace{-1mm}
\subsection{Ablation studies}

We conducted comprehensive ablation studies to evaluate the effectiveness of each component in MoRel, as summarized in Tab.~\ref{table:ablation_study}. All results were computed on a 300-frame subset sampled from the $\textbf{SelfCap}_{\text{LR}}$ sequences. 
Since MoRel independently trains each BDW (Fig.~\ref{fig:main_figure}, Sec.~\ref{sec:bidirectional deformation}) in a progressive manner, this subset evaluation reliably reflects the general trend of each component's influence. The baseline, variant (a), is a single-stage deformation model trained with a single $\mathbf{A}^{\text{Global}}$. This setup maintains global consistency but fails to capture detailed local motion, leading to the lowest quantitative results.
Variant (b) introduces KfAs that form local canonical spaces for each temporal segment, enhancing regional consistency and motion fidelity. This improvement in LPIPS support this claim. 
Moreover, this localized partitioning significantly reduces both training memory and rendering memory, thanks to the \textit{on-demand loading} in Alg.~\ref{alg:MoRel_training}-line 12 and 16. 
Next, variant (c) incorporates PWD and Linear Blending which linearly interpolates between adjacent KfAs based on temporal distance. 
Although the bidirectional deformation slightly increases the training memory but marginal, and we observe that even simple linear blending leads to noticeable quality improvements, by preventing inter-chunk interference. 
Replacing simple interpolation with IFB in variant (d) further improves the representation fidelity. 
This gain came from the learnable opacity control, which enables the model to effectively represent the scene even various forms of irregular motion.  
Finally, variant (e) applies FHD to balance between quality and memory usage. By efficiently control the anchor densification guided by feature-variance, we achieve nearly identical or slightly improved performance while simultaneously reducing the rendering memory usage. 
Consequently, MoRel demonstrates that the integration of the proposed components enables effective long-range 4D motion representation while maintaining bounded memory usage. Further diverse analyses are provided in \textit{Suppl.~\ref{Analysis and Ablations}}.

\begin{table}[t]
\small
\begin{center}
\setlength\tabcolsep{4pt} 
\renewcommand{\arraystretch}{1.4}
\scalebox{0.65}{
\begin{tabular}{ l | c c c | c c }
\bottomrule
\hline\noalign{\smallskip}
\multirow{2}{*}{Variant} & 
\multirow{2}{*}{PSNR$\uparrow$} & 
\multirow{2}{*}{SSIM$\uparrow$} & 
\multirow{2}{*}{LPIPS$\downarrow$} & 
\multicolumn{2}{c}{Memory (MB)$\downarrow$} \\  
\cline{5-6}\noalign{\smallskip}
& & & & Training & Rendering \\
\bottomrule
\hline\noalign{\smallskip}
(a) 2-stage, GCA-only (+) uni-deformation  & 19.71 & 0.654 & 0.386 & $\sim$12,000 & 156 \\
(b) 3-stage, (a) with KfA & 19.90 & 0.647 & \colorbox{GreenYellow}{0.364} & $\sim$4,500 & 94 \\
(c) 3-stage, (b) with PWD and Linear Blend & 20.66 & 0.656 & 0.358 & $\sim$6,500 & 138 \\
(d) 4-stage, (b) with PWD and IFB & \colorbox{GreenYellow}{21.07} & 0.672 & \colorbox{GreenYellow}{0.342} & $\sim$6,500 & 144 \\
(e) 4-stage, (d) with FHD & 21.20 & 0.672 & 0.348 & $\sim$6,000 & 126 \\
\bottomrule
\hline\noalign{\smallskip}
\end{tabular}}
\end{center}
\vspace{-0.5cm}
\caption{\textbf{Ablation studies on MoRel components.} Each row evaluates the impact of a specific design choice. Yellow-green cells highlight configurations with substantial gain.}
\label{table:ablation_study}
\end{table}

\section{Conclusion}
We present MoRel, an on-demand loading-based long-range 4D Gaussian Splatting framework that delivers flicker-free, memory-bounded, and faithful reconstructions of dynamic scenes. At its core, Anchor Relay–based Bidirectional Blending (ARBB) coordinates Key-frame Anchors (KfAs) and fuses their bidirectional deformations via learnable temporal opacity to maintain coherent motion. Feature-variance–guided Hierarchical Densification (FHD) further enriches anchor regions to recover high-frequency detail while avoiding overpopulation of anchor-points. Extensive experiments on SelfCap$_{\text{LR}}$ reveal that MoRel consistently attains sharper reconstructions, smoother temporal transitions and lower memory consumption, positioning it as a compelling and scalable solution for real-world long-range 4D motion modeling.
\label{Conclusion}
\section{Acknowledgement}
This work was supported by the IITP grant funded by the Korea government (MSIT): Development of immersive video spatial computing technology for ultra-realistic metaverse services (No.2022-0-00022, RS-2022-II220022), 
and by the National Research Foundation of Korea (NRF) grant funded by the Korea government (MSIT) (RS-2025-23524035), 
and was partly supported by Institute of Information \& communications Technology Planning \& Evaluation (IITP) grant funded by the Korea government (MSIT) (No. RS-2025-25422680, Metacognitive AGI Framework and its Applications, 25\%).

{
    \small
    \bibliographystyle{ieeenat_fullname}
    \bibliography{main}
}

\clearpage
\setcounter{page}{1}
\maketitlesupplementary
\appendix
\section{Notation}
\label{sec:notation}
\begin{figure}[!t]
  \centering
\includegraphics[scale=0.72]{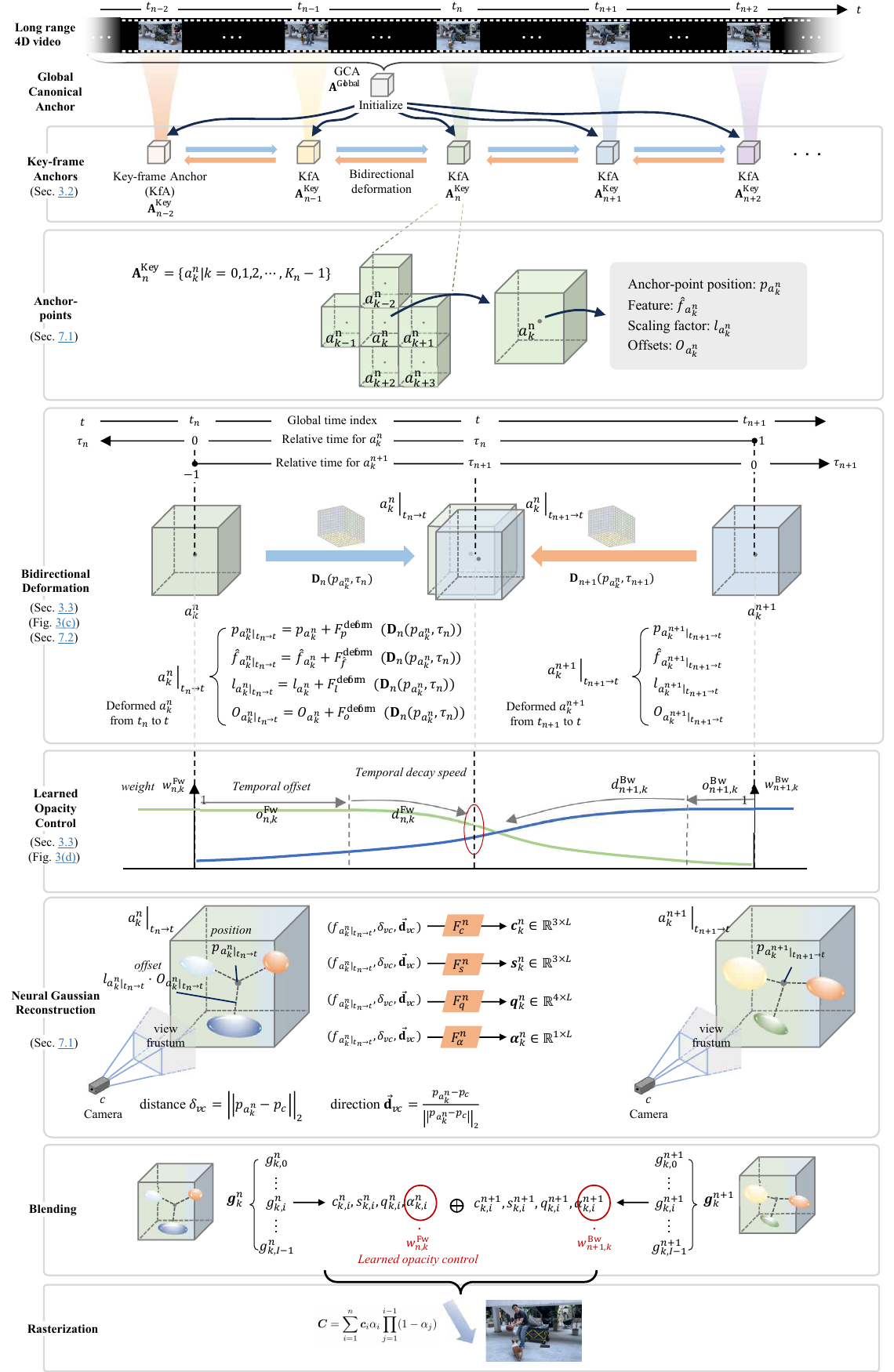}
\vspace{-1mm}
\caption{\textbf{Overview of processing pipeline including the relationship among notations.}}
\label{fig:Overview of processing pipeline}
\end{figure}

\vspace{-2mm}
In this paper, we use the following notations. An overview of our processing pipeline, illustrating the relationships among the notations, is provided in Fig.~\ref{fig:Overview of processing pipeline}.
\vspace{-2mm}

\vspace{1mm}
\begin{itemize}
  \item \textbf{Anchor}
    \begin{itemize}
      \item $\mathbf{A}$: Anchor (canonical) space representing a static scene.
      \item $\mathbf{A}^{\text{Key}}$: Key-frame anchor (KfA).
      \item $\mathbf{A}^{\text{Global}}$: Global Canonical Anchor (GCA).
      \item $\mathbf{\widetilde{A}}^{\text{Global}}$: Level-assigned GCA by FHD.
      \item $\mathbf{A}^{\text{blended}}_t$: Blended anchor by adjacent KfAs in IFB training stage.
      \item $n$: Index of KfA.
      \item $N$: Total number of KfAs. 
    \end{itemize}
\textit{e.g.,} $\mathbf{A}^{\text{Key}}_n$ denotes the $n$-th KfA.
  
\vspace{1mm}
  \item \textbf{Anchor-points}
    \begin{itemize}
      \item $a$: Anchor-points that constitute an anchor space $\mathbf{A}$, formed by discretizing $\mathbf{A}$ into a grid; each point has attributes defined in Sec.\ref{Preliminary}.
      \item $a^n$: Anchor-points belonging to the $n$-th KfA.
      \item $k$: Index of anchor-points.
      \item $K_n$: Total number of anchor-points in the $n$-th KfA.
    \end{itemize}
\textit{e.g.,} $\mathbf{A}^{\text{Key}}_n = \{\, a^{n}_k \mid k = 0, 1, \cdots, K_n - 1 \,\}.$
    
\vspace{1mm}
  \item \textbf{Neural Gaussians:} per–$a$ Gaussian set.
    \begin{itemize}
      \item $i$: Index of neural Gaussians assigned to an anchor-point.
      \item $I$: Number of neural Gaussians per anchor-point.  
    \end{itemize}
\textit{e.g.,} $I=10$ indicates each anchor-point has 10 neural Gaussians. Note that $I$ is not a parameter dependent on $n$ or $k$; it is a constant shared across all KfAs and their anchor-points.

\vspace{1mm}
  \item \textbf{Temporal indices}
    \begin{itemize}
      \item $t$: Temporal index (frame).
      \item $T$: Total number of frames in the video sequence.
      \item $t_n$: Frame index corresponding to $\mathbf{A}^{\text{Key}}_n$. 
    \end{itemize}
 
\vspace{1mm}
  \item \textbf{Deformation field}
    \begin{itemize}
      \item $\mathbf{D}_n(\cdot, \tau)$: Deformation field of the $n$-th KfA that warps anchor-points according to the normalized relative time $\tau \in [-1,1]$.  
            A single deformation field models both forward (+) and backward (-) temporally bidirectional directions in a unified manner.
      \item $\tau$: Normalized relative time between neighboring KfAs; $\tau = -1$ and $\tau = 1$ correspond to backward and forward endpoints, respectively. 
      \item $\tau_n$: Normalized relative time for $A^{key}_n$.
    \end{itemize}
    \textit{e.g.,} $\mathbf{D}_n(a^{n}_k, \tau_n)$ represents the temporally deformed amount for attributes of anchor-point $a^{n}_k$ at $\tau_n$, where $\tau_n\in[-1,1]$ is a relative time corresponding to $t\in[t_{n-1},t_{n+1}]$.

\vspace{1mm}
  \item \textbf{Blending parameters}
  \begin{itemize}
    \item $\textbf{o}_{n}^{\text{dir}}$: A set of temporal offsets assigned to $\textbf{A}_n^{\text{Key}}$
     \item $\textbf{d}_{n}^{\text{dir}}$: A set of temporal decay speeds assigned to $\textbf{A}_n^{\text{Key}}$
    \item $o_{n,k}^{\text{dir}}$: Temporal offset assigned to $a^n_k$, $\text{dir}\in\{{\text{Fw, Bw}}\}$
    \item $d_{n,k}^{\text{dir}}$: Temporal decay speed assigned to $a^n_k$.     
    \item $w_{n,k}^{\text{dir}}$: Temporal blending weight assigned to $a^n_k$, derived by Eq.~\ref{eq:opacity control}.
  \end{itemize}

\vspace{1mm}
  \item \textbf{Initial points and cameras}
  \begin{itemize}
    \item $\mathcal{P}$: Input point cloud for initializing $\textbf{A}^{\text{Global}}$.
    \item $\textbf{C}$: A set of input training camera frames. 
    \item $m$: Index of camera.
    \item $M$: Total number of cameras. 
    \item $c_t^m$: an individual frame of $m$-th camera at time $t$. 
  \end{itemize}
  \textit{e.g.,} $\textbf{C}=\{c_t^m | t=0,\cdots,T-1, m=0,\cdots,M-1\}$

\vspace{1mm}
\item \textbf{Training index and stages}
  \begin{itemize}
    \item $j^\mathcal{S}$: Training iteration index within one of the four training stages, 
          $\mathcal{S} \in \{\text{GCA}, \text{KfA}, \text{PWD}, \text{IFB}\}$.
    \item $j^\mathcal{S}_n$: Training iteration index for $n$-th KfA, within the training stage $\mathcal{S}$.\\
    \textit{e.g.,} $j^{\text{KfA}}_n$ denotes a training index for $\mathbf{A}^{\text{KfA}}_n$ in KfA training stage. 
    \item $J^\mathcal{S}$: Total training iteration of the stage $\mathcal{S}$. 
    \item $J^\mathcal{S}_n$: Total training iteration of $\mathbf{A}^{\text{KfA}}_n$ for the stage $\mathcal{S}$. \\
    \textit{e.g.,} $J^\text{PWD} = \sum_{n=0}^{N-1} J^{\text{PWD}}_n$. 
    \item Note that, in the GCA training stage, training is executed once to initialize $\mathbf{A}^{\text{Global}}$.
  \end{itemize}

\vspace{1mm}
  \item \textbf{Temporal units and windows}
    \begin{itemize}
      \item $\mathrm{GOP}$ (Group of Pictures): Temporal spacing between adjacent KfAs. \textit{e.g.,} $\mathrm{GOP}=100$ means each KfA is located at every 100 frames. \\
       $\forall{n}\in[1, N-1], \quad t_n-t_{n-1}=\text{GOP}$
      \item Chunk: Temporal range that can be rendered without reloading new KfA, equivalent to the temporal range covered by a unidirectional deformation of a KfA.
       \item $\text{Chunk}_{n}$: $n$-th chunk that has temporal range of $[\,t_n,\, t_n + \mathrm{GOP}\,]$.
      \item BDW (Bidirectional Deformation Window): Temporal range modeled by a single KfA via bidirectional deformation. 
      \item $\text{BDW}_n$ : $n$-th BDW that has temporal range of $[\,\max(0,\, t_n - \mathrm{GOP}),\, \min(t_n + \mathrm{GOP},\, T-1)\,]$.
      \item $\epsilon$: Temporal tolerance for robust training of KfA. 
    \end{itemize}
\end{itemize}

\clearpage
\onecolumn
\clearpage
\twocolumn

\section{Preliminary}
\label{Preliminary}
\subsection{Anchor-point-based Representation}
\label{Preliminary:Scaffold}
We adopt an anchor-point-based scene representation first introduced in Scaffold-GS~\cite{lu2024scaffold} with sparse point cloud initialization~\cite{schonberger2016structure}.
We denote a set of anchor-point positions by $\textbf{A} = \{a_k\}_{k=0}^{K-1}$, where the $k$-th anchor-point is associated with a learnable attribute $\theta_k = (p_k, \hat{f}_k, \ell_k, O_k)$ as shown in the Anchor-points stage of Fig.~\ref{fig:Overview of processing pipeline}, 
where $p_k \in \mathbb{R}^3$ is the position of the anchor-point, $\hat{f}_k$ is an anchor-point feature that summarizes the local geometry and appearance around $a_k$, 
$\ell_k \in \mathbb{R}^3$ is a scaling vector controlling the spatial extent of Gaussians associated with the anchor-point, 
and $O_k \in \mathbb{R}^{I \times 3}$ is a set of relative offsets that specify the canonical positions of the Gaussians attached to anchor-point $k$. 
Let $\delta_{k,\text{cam}}$ denote the 3D displacement from the camera center to anchor-point $a_k$, and 
\(\overrightarrow{\mathbf{d}}_{k,\text{cam}}\) the viewing direction. 
The attributes of the neural Gaussians associated with anchor-point $k$ are then defined as
\begin{equation}
    \{\textit{attr}_{k,i}\}_{i=0}^{I-1}
    =
    F_{\textit{attr}}(\hat{f}_k,\; \delta_{k,\text{cam}},\; \overrightarrow{\mathbf{d}}_{k,\text{cam}}),
    \label{eq:anchor_attributes}
\end{equation}
where each Gaussian attribute $\textit{attr}_{k,i}$ contains color \(c_{k,i}\), 
quaternion orientation \(q_{k,i}\), scale \(s_{k,i}\) and opacity \(\alpha_{k,i}\) as illustrated in Neural Gaussian Reconstruction stage in Fig.~\ref{fig:Overview of processing pipeline}. Note that the position of each Gaussian is directly obtained by inner product of $\ell_k$ and $O_k$.

To adapt anchor-points to the local complexity of a scene, the anchor-points undergo densification, i.e., growing and pruning. The densification process is controlled by three parameters: (i) a success threshold that measures how many views have observed the anchor-point, (ii) an opacity threshold that determines how visible an anchor-point is, and (iii) a gradient threshold that identifies anchor-points requiring further adjustment (i.e., candidates for densification). Among these, the gradient criterion plays the most critical role. 
Within certain intervals, each anchor-point $a_k$ accumulates gradients $g_{G_{k,i}}$, and this value is compared against the preset gradient threshold to decide whether densification should occur. 
Hence, anchor-points with larger accumulated gradients are prioritized during densification.
Our FHD (Sec.~\ref{sec:hierarchical densification}) further modulates this process by applying frequency-aware weighting to the accumulated gradients, enabling more controlled and adaptive densification.


\begin{table*}[t]
\begin{center}
\setlength\tabcolsep{3pt}
\renewcommand{\arraystretch}{1.25}
\scalebox{0.8}{%
\begin{tabular}{ c | c | c | c | c | c | c | c }
\toprule
Dataset & Scene & \#Images & FPS & \#Cameras & Training resolution & Average OFps & Normalized camera distance \\
\midrule
\multirow{5}{*}{SelfCap$_{\text{LR}}$} 
  & Bike$_1$ & 3600 & 60 & 22 & $1080\times1080$ & 15.55 & 0.109 \\
  & Bike$_2$ & 3600 & 60 & 22 & $1080\times1080$ & 16.40 & 0.109 \\
  & Corgi & 3400 & 60 & 24 & $1920\times1080$ & 72.96 & 0.079 \\
  & Yoga & 3600 & 60 & 24 & $1920\times1080$ & 36.74 & 0.155 \\
  & Dance & 3600 & 60 & 24 & $1920\times1080$ & 79.91 & 0.112 \\
\midrule
\multirow{6}{*}{PanopticSports} 
  & Basketball & 150 & 30 & 31 & $640\times360$ & 27.29 & 0.073 \\
  & Boxes      & 150 & 30 & 31 & $640\times360$ & 26.17 & 0.091 \\
  & Football   & 150 & 30 & 31 & $640\times360$ & 25.92 & 0.085 \\
  & Juggle     & 150 & 30 & 31 & $640\times360$ & 24.55 & 0.075 \\
  & Softball   & 150 & 30 & 31 & $640\times360$ & 23.91 & 0.101 \\
  & Tennis     & 150 & 30 & 31 & $640\times360$ & 21.53 & 0.088 \\
\midrule
\multirow{7}{*}{DyCheck-iPhone} 
  & Apple & 475 & 30 & 3 & $360\times480$ & 3.79 & 0.001 \\
  & Block & 350 & 30 & 3 & $360\times480$ & 30.42 & 0.001 \\
  & Paper-windmill & 277 & 30 & 3 & $360\times480$ & 23.19 & 0.001 \\
  & Space-out & 429 & 30 & 3 & $360\times480$ & 6.19 & 0.001 \\
  & Spin & 426 & 30 & 3 & $360\times480$ & 38.35 & 0.001 \\
  & Teddy & 350 & 30 & 3 & $360\times480$ & 11.10 & 0.001 \\
  & Wheel  & 250 & 30 & 3 & $360\times480$ & 36.26 & 0.001 \\
\bottomrule
\end{tabular}
} 
\vspace{2mm}
\caption{\textbf{Dataset statistics.} For each scene of the three datasets, we summarize the number of images, Frames-Per-Second (FPS), the number of cameras, the training image resolution, the average OFps, and the normalized camera distance used in our experiments.} 
\label{tab:dataset_scene_statistics}
\end{center}
\end{table*}

\subsection{Hexplane Deformation}
\label{Preliminary:Hexplane}
For modeling temporal motion, we adopt a hexplane-based deformation field that warps anchor-points according to a normalized relative time variable, extending the deformation field introduced in~\cite{wu20244d} to operate \textit{bidirectionally}. 
For each $\textbf{A}^{\text{Key}}_n$, we define a bidirectional deformation field $\mathbf{D}_n(a^{n}_k, \tau_n)$, where $\tau_n \in [-1,1]$ denotes the bidirectional temporal coordinate that corresponds to $[t_{n-1}, t_{n+1}]$.
Given an anchor-point $a^n_k$ belonging to $\mathbf{A}^{\text{Key}}_n$, $\mathbf{D}_n(a^{n}_k, \tau_n)$ encodes the temporally warped attributes of $a^n_k$ at relative time $\tau_n$
as shown in Bidirectional Deformation stage in Fig.~\ref{fig:Overview of processing pipeline}.

\begin{figure*}[h]
\begin{center}
\includegraphics[scale=0.62]{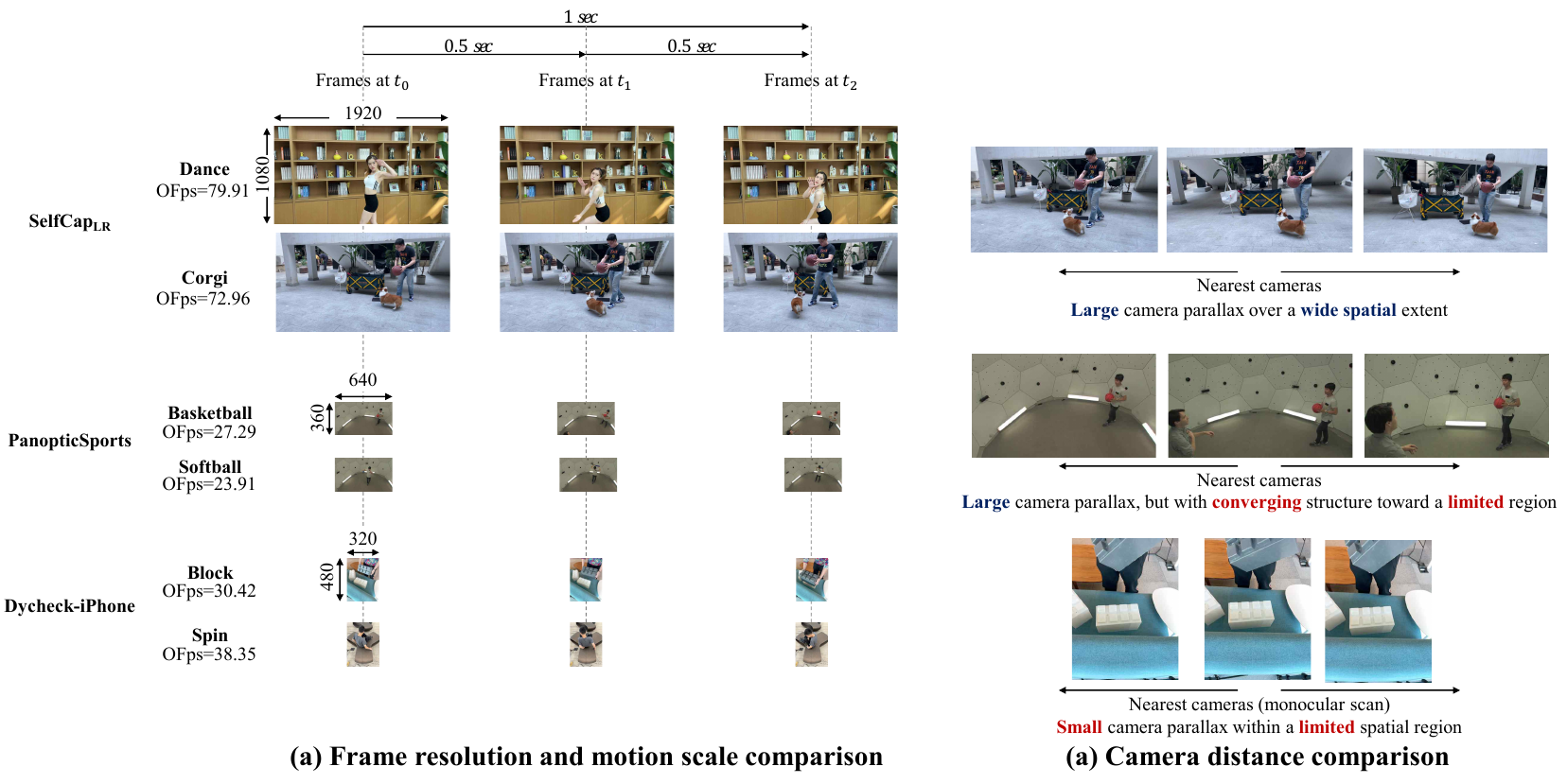}
\vspace{2mm}
\caption{\textbf{Dataset visualization.} For the datasets analyzed in Tab.~\ref{tab:dataset_scene_statistics} and Sec.~\ref{Dataset Statistics}, (a) we visualize frames sampled every 0.5 seconds over a 1-second duration. The size of each frame is scaled according to its relative resolution, and OFps shown below each sequence name indicates the average optical flow magnitude per second. We also show (b) the distance between the closest pair of cameras. As can be seen from the visualizations, $\textbf{SelfCap}_{\text{LR}}$ exhibits high resolution, fast motion within a unit time, and a large camera parallax on wide spatial extent.}
\label{fig:dataset}
\end{center}
\end{figure*}

\subsection{tOF Metric}
\label{tOF}
Previous works~\cite{kwak2025modec, park2025splinegs} use the tOF metric~\cite{chu2020learning} to evaluate temporal consistency. This metric measures the discrepancy between motion fields estimated from consecutive frames of the reference sequence and those of the rendered sequence. Specifically, optical flow is computed between two consecutive frames for both sequences, and the pixel-wise difference between the two motion fields is measured. The tOF metric is defined as

\begin{equation}
tOF = \|OF(b_{t-1}, b_t) - OF(g_{t-1}, g_t)\|_1 ,
\end{equation}
where $OF(\cdot)$ denotes the optical flow estimation operator, $b_t$ represents the reference frame, and $g_t$ denotes the rendered frame. By comparing motion fields rather than raw pixel values, this metric evaluates temporal motion consistency between consecutive frames.

\section{Dataset}
\label{Dataset analysis}
\subsection{Dataset Statistics}
\label{Dataset Statistics}

We construct the $\textbf{SelfCap}_{\text{LR}}$ dataset to enable a rigorous evaluation of long-range motion, by reorganizing the original SelfCap~\cite{wang2025freetimegs,xu2024longvolcap} sequences into long-duration captures with wide-baseline multi-view observations. To validate the statistical properties and difficulty of $\textbf{SelfCap}_{\text{LR}}$, we compare it with established large-motion benchmarks, PanopticSports~\cite{luiten2024dynamic} and DyCheck-iPhone~\cite{Gao2022Dycheck}, under a unified evaluation protocol, as summarized in Tab.~\ref{tab:dataset_scene_statistics}.

The table reports the number of images, FPS, number of cameras, resolution, average Optical Flow magnitude per second (OFps), and normalized camera distance. 
Specifically, to quantify motion magnitude, we compute OFps by estimating optical flow using RAFT~\cite{teed2020raft} over 1-second temporal windows from a fixed test camera. 
For each window, we retain only pixels whose flow magnitude exceeds a small threshold to isolate dynamic regions, and then average the $\ell_2$ norm of the remaining flow vectors.
Camera distance is defined as follows. We obtain camera centers from COLMAP~\cite{schonberger2016structure} and, for each camera, compute the Euclidean distance to its nearest-neighbor camera center. To remove dataset-specific scale effects, we estimate the scene size from the COLMAP-reconstructed 3D point cloud. Specifically, we build an axis-aligned bounding box by taking the minimum and maximum coordinates along the x, y, and z axes, and use the diagonal length of this box as a representative scene length. The normalized camera distance is then given by the nearest-neighbor camera distance divided by this bounding-box diagonal. This normalization enables scale-invariant comparison of relative camera spacing and baselines across datasets, preventing absolute scene size from dominating the distance measure. 

In addition, Fig.~\ref{fig:dataset} provides visual support for the quantitative results presented in Tab.~\ref{tab:dataset_scene_statistics}. 
As shown in Fig.~\ref{fig:dataset}(a), the $\textbf{SelfCap}_{\text{LR}}$ exhibits significantly higher spatial resolution and much larger motion magnitude per unit time (1 second) compared to other datasets. 
Fig.~\ref{fig:dataset}(b) further visualizes comparison of the nearest camera views. 
Our $\textbf{SelfCap}_{\text{LR}}$ features strong camera parallax across a wide spatial extent, whereas PanopticSports contains large parallax but from inward-facing cameras converging toward a limited region, and DyCheck-iPhone consists of low-parallax monocular captures within a much more limited space. These characteristics, combined with the long-range temporal range (around 3.5k frames at 60 FPS) of $\textbf{SelfCap}_{\text{LR}}$, make motion modeling of this dataset particularly challenging.

\subsection{Initial Point Cloud}
\label{initial point cloud}

Using temporally dense point clouds \cite{sun20243dgstream, wang2025freetimegs,li2025gifstream4dgaussianbasedimmersive} imposes a substantial burden on training for long-range video and becomes a practical hurdle for real-world applications. 
Motivated by this, we use a single set of point clouds to initialize MoRel. This point cloud is constructed by merging point clouds sampled thousands of frames apart along the temporal axis and repeatedly applying voxel-based decimation (voxel size = 0.01) until the total number of points is reduced to below 60k. 
The resulting sparse point cloud is used to initialize $\textbf{A}^{\text{Global}}$ in the GCA training stage \ref{sec:anchor relay}. For fair comparison, we apply the same constructed point cloud set to all baseline models, except for \cite{sun20243dgstream}. 
\begin{itemize}
    \item Bike$_1$ and Bike$_2$: 2 point clouds sampled at 2000-frame intervals are merged. 
    \item Corgi, Yoga, and Dance: 4 point clouds sampled at 1000-frame intervals are merged. 
\end{itemize}
For \cite{sun20243dgstream}, since it requires a GOP-based point cloud initialization, we followed its original protocol on the $\textbf{SelfCap}_{\text{LR}}$.

\section{Additional Results}
\label{Additional Results}
\subsection{Standard Benchmark Experiments and Generalization Ability}
To validate the generalization capability of our proposed method, we conducted evaluations on multiple standard benchmark datasets with diverse environmental characteristics. The experimental results for the Dycheck~\cite{Gao2022Dycheck} dataset are summarized in Tab.~\ref{table:iphone_comparison}, while additional results for HyperNeRF~\cite{park2021hypernerf}, DyNeRF~\cite{li2022neural}, and PanopticSports~\cite{luiten2024dynamic} are presented in Tab.~\ref{tab:quant_tables}. All baseline results are sourced from MoDec-GS~\cite{kwak2025modec} and GIFStream~\cite{li2025gifstream4dgaussianbasedimmersive}. To ensure a fair comparison, we strictly adhered to the experimental protocols and settings established in these prior works.

As illustrated in Tab.~\ref{tab:dataset_scene_statistics} and Fig.~\ref{fig:dataset}, the benchmark datasets, including Dycheck, exhibit fundamental characteristics that differ significantly from $\textbf{SelfCap}_{\text{LR}}$ in terms of capture distance, viewpoint configuration, and motion range. Notably, even in relatively short-video sequences that do not involve long-range motions, our method achieves robust reconstruction performance comparable to SOTA approaches without any dataset-specific tuning.

During experimentation, we maintained most of the core hyperparameters optimized for $\textbf{SelfCap}_{\text{LR}}$, such as densification parameters, while only minimally adjusting the GOP and hexplane sizes to accommodate the specific sequence lengths. Although we report results from this naïve extension due to time constraints, these findings demonstrate that our method maintains strong reconstruction quality and efficient model capacity across various standard environments. This indicates that our model possesses the versatility to be immediately applied to diverse scenarios without the need for excessive parameter engineering.

\subsection{Demo Video}
The goal of our framework is to consistently and finely represent long-range 4D motion without underfitting or temporal flickering (Fig.~\ref{fig:figure_page1}). Such properties cannot be fully validated by numerical metrics alone; they must also be examined through actual rendered video results. 
To this end, we provide a \href{https://www.youtube.com/watch?v=r8KNhreRf3Q}{demo video} of novel view synthesis results on the SelfCap$_\text{LR}$ dataset. We highly encourage readers to view it. 

\begin{table*}[htbp]
\begin{center}
\setlength\tabcolsep{5pt}
\renewcommand{\arraystretch}{1.1}

\scalebox{0.8}{
\begin{tabular}{ c |ccc ccc ccc ccc }
\bottomrule \hline
\noalign{\smallskip}
Method 
& \multicolumn{3}{c}{Apple}
& \multicolumn{3}{c}{Block}
& \multicolumn{3}{c}{Paper-windmill}
& \multicolumn{3}{c}{Space-out} 
\\  
\hline\noalign{\smallskip}
& mPSNR & mSSIM & Storage
& mPSNR & mSSIM & Storage
& mPSNR & mSSIM & Storage
& mPSNR & mSSIM & Storage
\\
\hline

SC-GS~\cite{Huang2024SCGS}    
& 14.96 & 0.692 & 173.3
& 13.98 & 0.548 & 115.7
& 14.87 & \best{0.221} & 446.3
& \second{14.79} & 0.511 & 114.2 
\\

Deformable 3DGS~\cite{Yang2024Deformable3DGS}       
& 15.61 & 0.696 & 87.71
& 14.87 & 0.559 & 118.9
& 14.89 & 0.213 & 160.2
& 14.59 & 0.510 & \second{42.01}
\\

4DGS~\cite{wu20244d}   
& 15.41 & 0.691 & 61.52
& 13.89 & 0.550 & 63.52
& 14.44 & 0.201 & 123.9
& 14.29 & \second{0.515} & 52.02
\\

MoDec-GS~\cite{kwak2025modec}
& \second{16.48} & \second{0.699} & \best{23.78}
& \best{15.57} & \best{0.590} & \best{13.65}
& \second{14.92} & \second{0.220} & \best{17.08}
& 14.65 & \best{0.522} & \best{18.24}
\\

\textbf{MoRel (Ours)}
& \best{16.92} & \best{0.702} & \second{38.37}
& \second{15.12} & \second{0.572} & \second{52.81}
& \best{14.96} & 0.213 & \second{63.92}
& \best{15.30} & 0.519 & 79.55
\\

\bottomrule \hline
\end{tabular}
}

\scalebox{0.8}{
\begin{tabular}{ c |ccc ccc ccc |ccc }
\hline\noalign{\smallskip}
& \multicolumn{3}{c}{Spin}
& \multicolumn{3}{c}{Teddy}
& \multicolumn{3}{c}{Wheel}
& \multicolumn{3}{c}{\textbf{Average}}
\\  
\hline\noalign{\smallskip}
& mPSNR & mSSIM & Storage
& mPSNR & mSSIM & Storage
& mPSNR & mSSIM & Storage
& mPSNR & mSSIM & Storage
\\
\hline

SC-GS~\cite{Huang2024SCGS}    
& 14.32 & 0.407 & 219.1
& 12.51 & 0.516 & 318.7
& \second{11.90} & \second{0.354} & 239.2
& 13.90 & 0.464 & 232.4 
\\

Deformable 3DGS~\cite{Yang2024Deformable3DGS}    
& 13.10 & 0.392 & 133.9
& 11.20 & 0.508 & 117.1
& 11.79 & 0.345 & 106.1
& 13.72 & 0.461 & 109.4
\\

4DGS~\cite{wu20244d}  
& 14.89 & 0.413 & 71.80
& 12.31 & 0.509 & 80.44
& 10.83 & 0.339 & 96.50
& 13.72 & 0.460 & 78.54
\\

MoDec-GS~\cite{kwak2025modec}
& \second{15.53} & \second{0.433} & \best{26.84}
& \second{12.56} & \second{0.521} & \best{12.28}
& \best{12.44} & \best{0.374} & \best{16.68}
& \second{14.60} & \best{0.480} & \best{18.37}
\\

\textbf{MoRel (Ours)}
& \best{15.70} & \best{0.448} & \second{68.56}
& \best{13.14} & \best{0.522} & \second{40.59}
& 11.77 & 0.350 & \second{78.76}
& \best{14.70} & \second{0.475} & \second{62.63}
\\

\bottomrule \hline
\end{tabular}
}
\caption{\textbf{Quantitative results comparison on the DyCheck-iPhone dataset \cite{Gao2022Dycheck}.} 
Each block element denotes (mPSNR(dB)$\uparrow$ / mSSIM$\uparrow$ /  Storage(MB)$\downarrow$).}
\label{table:iphone_comparison}
\end{center}
\end{table*}

\begin{table*}[t]
\centering
\renewcommand{\arraystretch}{1.05}
\begin{subfigure}[t]{0.29\textwidth}
\centering
\resizebox{\linewidth}{!}{%
\begin{tabular}{l c}
\hline
\multicolumn{2}{c}{(a) HyperNeRF} \\
\hline
Methods & PSNR/SSIM/LPIPS \\
\hline
4DGS & 27.24 / 0.793 / 0.285 \\
MoDec-GS & \second{27.37} / \second{0.811} / \best{0.232} \\
\hline
\textbf{MoRel} & \best{27.40} / \best{0.819} / \second{0.235} \\
\hline
\end{tabular}}
\end{subfigure}\hfill
\begin{subfigure}[t]{0.29\textwidth}
\centering
\resizebox{\linewidth}{!}{%
\begin{tabular}{l c}
\hline
\multicolumn{2}{c}{(b) DyNeRF} \\
\hline
Methods & PSNR/SSIM/LPIPS \\
\hline
MoDec-GS & 30.87 / 0.933 / \second{0.049} \\
GIFStream & \best{31.75} / \second{0.938} / 0.051 \\
\hline
\textbf{MoRel} & \second{31.06} / \best{0.946} / \best{0.046} \\
\hline
\end{tabular}}
\end{subfigure}\hfill
\begin{subfigure}[t]{0.29\textwidth}
\centering
\resizebox{\linewidth}{!}{%
\begin{tabular}{l c}
\hline
\multicolumn{2}{c}{(c) PanopticSports} \\
\hline
Methods & PSNR/SSIM/LPIPS \\
\hline
4DGS & 27.22 / 0.910 / \best{0.100} \\
MoDec-GS & \second{27.96} / \best{0.950} / 0.130 \\
\hline
\textbf{MoRel} & \best{28.21} / \second{0.930} / \second{0.117} \\
\hline
\end{tabular}}
\end{subfigure}
\caption{\textbf{Quantitative comparison on three datasets}. We report results on HyperNeRF, DyNeRF, and PanopticSports. Each block element of 3-performance denotes (PSNR(dB)$\uparrow$ / SSIM$\uparrow$ / LPIPS$\downarrow$).}
\label{tab:quant_tables}
\end{table*}

\section{Further Analysis and Ablations}
\label{Analysis and Ablations}
\subsection{Overall Training Time}
\label{Overall Training Time}
Although our method adopts a multi-stage training pipeline, increasing the number of stages does not necessarily translate into an excessive runtime overhead.
Tab.~\ref{tab:training_config} compares 4DGS~\cite{wu20244d} and MoRel on the Bike$_1$ and Yoga sequences (3.6k frames), reporting stage-wise iterations, iterations per second (it/s), and the total training time.
While MoRel requires more iterations in total than 4DGS, (i) the smaller hexplane configuration and (ii) the shorter observation frame window keep the per-stage throughput high.
Consequently, the overall training time does not grow proportionally to the increased iteration count and remains within a reasonable range in practice, indicating that the performance gains are achieved without incurring impractical computational cost.
\begin{table*}[t]
\centering
\setlength{\tabcolsep}{2pt}
\renewcommand{\arraystretch}{1.2}
\footnotesize
\scalebox{1.2}{
\begin{tabular}{c |c | c c c | c c c c c | c c c c c}
\bottomrule
\multicolumn{2}{c|}{}
& \multicolumn{3}{c|}{\textbf{4DGS} (All-at-once)}
& \multicolumn{5}{c|}{\textbf{MoRel} (GOP=100)}
& \multicolumn{5}{c}{\textbf{MoRel} (GOP=200)} \\
\hline
\multicolumn{2}{c|}{Hexplane size}
& \multicolumn{3}{c|}{[64, 64, 64, 1800]}
& \multicolumn{5}{c|}{[64, 64, 64, 100]}
& \multicolumn{5}{c}{[64, 64, 64, 200]} \\
\hline
\multicolumn{2}{c|}{Num.GOP}
& \multicolumn{3}{c|}{1}
& \multicolumn{5}{c|}{36}
& \multicolumn{5}{c}{18} \\
\bottomrule
\multicolumn{2}{c|}{}
&  Coarse & Fine &  Total
& GCA & KfA & PWD &  IFB &  Total
&  GCA &  KfA &  PWD & IFB & Total \\
\hline
\multicolumn{2}{c|}{Iteration}
& 5k & 200k & 205k
& 3k & 5k$\times$37 & 15k$\times$36 & 5k$\times$36 & 908k
& 3k & 5k$\times$19 & 15k$\times$18 & 5k$\times$18 & 583k \\
\hline
\multirow{2}{*}{it/s} & Bike\_1
& 10 & 6 & -
& 11 & 22 & 15 & 15 & -
& 11 & 20 & 13 & 13 & - \\
\cline{2-15}
& Yoga
& 7 & 4 & -
& 6 & 17 & 11 & 11 & -
& 7 & 16 & 9 & 9 & - \\
\hline
\multirow{2}{*}{Overall Time} & Bike\_1
& \multicolumn{3}{c|}{\textbf{10h}}
& \multicolumn{5}{c|}{\textbf{16.5h}}
& \multicolumn{5}{c}{\textbf{12h}} \\
\cline{2-15}
& Yoga
& \multicolumn{3}{c|}{\textbf{14.5h}}
& \multicolumn{5}{c|}{\textbf{23h}}
& \multicolumn{5}{c}{\textbf{17h}} \\
\bottomrule
\multicolumn{2}{c|}{}
& \multicolumn{3}{c|}{ PSNR / SSIM / LPIPS}
& \multicolumn{5}{c|}{ PSNR / SSIM / LPIPS}
& \multicolumn{5}{c}{ PSNR / SSIM / LPIPS} \\
\hline

\multirow{2}{*}{Performance} & Bike\_1
& \multicolumn{3}{c|}{21.67 / 0.662 / 0.375}
& \multicolumn{5}{c|}{\second{22.32} / \best{0.668} / \best{0.321}}
& \multicolumn{5}{c}{\best{22.38} / \second{0.663} / \second{0.330}} \\
\cline{2-15}
& Yoga
& \multicolumn{3}{c|}{13.72 / 0.715 / 0.402}
& \multicolumn{5}{c|}{\best{22.18} / \best{0.780} / \best{0.297}}
& \multicolumn{5}{c}{\second{21.77} / \second{0.769} / \second{0.298}} \\
\bottomrule
\end{tabular}}
\caption{\textbf{Training efficiency of 4DGS and MoRel on Bike$_1$ and Yoga (3.6k frames)}. We report stage-wise iterations, iterations per second (it/s), and total training time. Each block element of 6-performance denotes (Iterations / it/s / Training Time(h)$\downarrow$ / PSNR(dB)$\uparrow$ / SSIM$\uparrow$ / LPIPS$\downarrow$).}
\label{tab:training_config}
\end{table*}

\subsection{GOP Size}
GOP is a system-level control parameter rather than a hyperparameter for performance tuning, as it governs requirements such as random accessibility and rate--distortion trade-offs. As shown in Tab.~\ref{tab:gop_hexplane}, we evaluate the full dataset by varying GOP and hexplane size pairs. Overall reconstruction performance (PSNR/SSIM/LPIPS) remains largely stable across configurations, while larger GOPs reduce the number of GOPs and yield a substantial reduction in total storage. These results indicate that GOP can be selected based on target system constraints (e.g., storage budget and access granularity) without incurring a significant performance drop.

\begin{table}[t]
\centering
\renewcommand{\arraystretch}{1.1}
\scalebox{0.6}{
\begin{tabular}{c | c | c c c | c | c}
\hline
GOP size & Hexplane & PSNR$\uparrow$ & SSIM$\uparrow$ & LPIPS$\downarrow$ & Num. GOP & Total storage (MB)$\downarrow$ \\
\hline
100 & {[}64,64,64,100{]} & \best{21.1478} & \best{0.670} & \best{0.336} & 12 & 862.6 \\
200 & {[}64,64,64,200{]} & 21.0281 & \second{0.667} & \second{0.344} & 6  & \second{641.8} \\
300 & {[}64,64,64,300{]} & \second{21.0741} & 0.662 & 0.345 & 4  & \best{552.6} \\
\hline
\end{tabular}}
\caption{\textbf{Comparison of reconstruction quality and storage efficiency across GOP sizes on $\textbf{SelfCap}_{\text{LR}}$-M}. We vary the GOP size and the corresponding hexplane configuration. Each block element of 5-performance denotes (PSNR(dB)$\uparrow$ / SSIM$\uparrow$ / LPIPS$\downarrow$ / Num.\ GOP / Storage(MB)$\downarrow$).}
\label{tab:gop_hexplane}
\end{table}

\subsection{FHD}
\label{Analysis on FHD}

\begin{figure*}[htbp]
\centering
\includegraphics[scale=0.7]{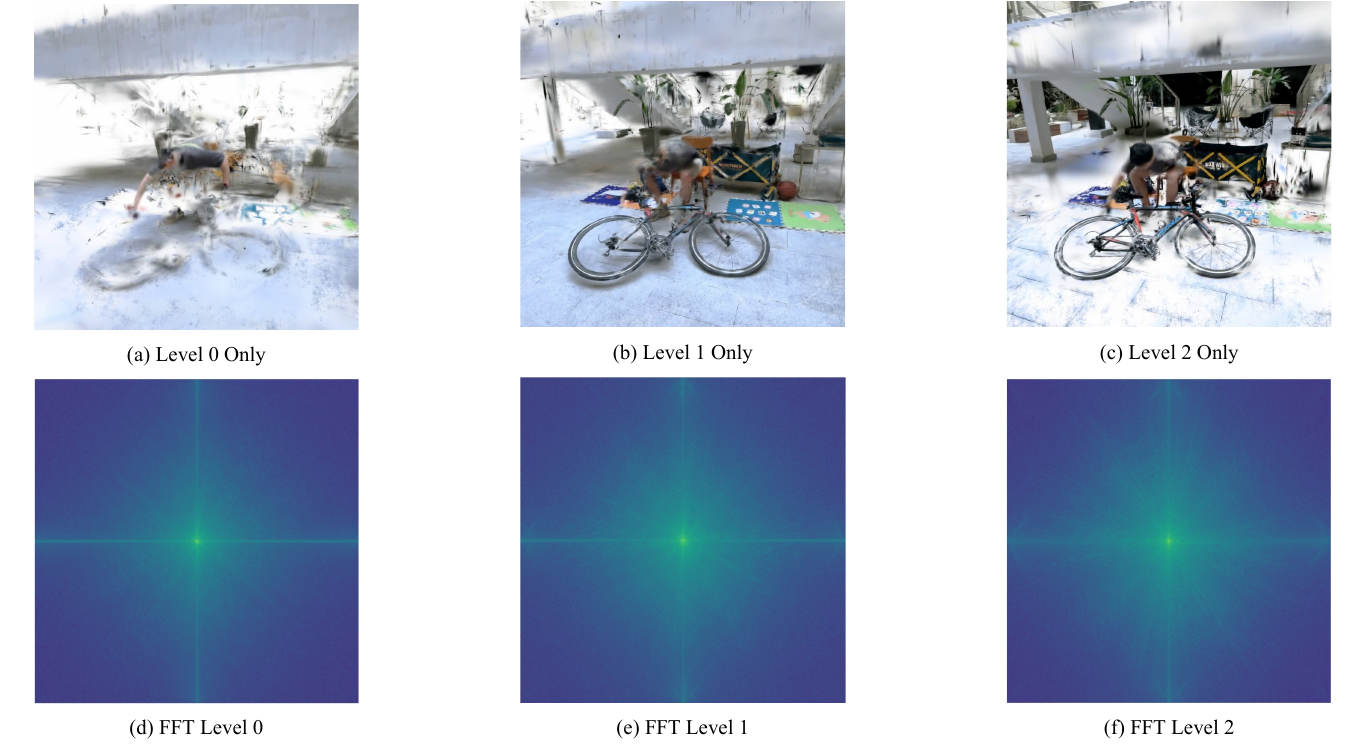}
\caption{\textbf{FHD Visualization and Frequency Analysis}.
(a–c) Level-specific renderings that retain only Gaussians associated with Level 0–2 anchor-points under FHD. (d–f) 2D FFT magnitudes of the corresponding renderings, showing increasingly dominant high-frequency components at higher levels.}
\label{fig:analyze}
\end{figure*}
We provide a comprehensive analysis of the hierarchical design and hyperparameters of FHD on $\textbf{SelfCap}_{\text{LR}}$-S (300 frames). As shown in Tab.~\ref{table:FHD_comparison}, increasing the number of hierarchy levels consistently improves reconstruction quality while reducing storage. With the default three-level setting ($L \in \{0,1,2\}$), the storage per $\mathbf{A}^{\text{Key}}_n$ drops from 72.9~MB to 57.3~MB, corresponding to a 21.4\% reduction. This gain is more pronounced in challenging scenes where accurate capacity allocation is critical, particularly those with large OFps such as Corgi (72.96) and Dance (79.91). In contrast, the two-level variant reduces storage but can slightly degrade quality in several sequences. This behavior is expected because a binary variance partition is too coarse to represent the continuous spectrum of scene complexity, forcing regions with intermediate texture or motion into either the low- or high-level group. Under the progressive densification schedule, such mis-grouping delays refinement in these regions and can lead to mild under-densification.

As shown in Tab.~\ref{tab:hyperparam_ablation}, we further analyze hyperparameter sensitivity by varying the variance thresholds $[\tau_1,\tau_2]$ in Eq.~\ref{eq:level assignment} and the level-wise weights $\lambda_L$ ($L \in \{0,1,2\}$) in Eq.~\ref{eq:densification}. Across a broad range of settings, performance remains nearly unchanged while the storage per $\mathbf{A}^{\text{Key}}_n$ can be reduced meaningfully (the first row denotes the default setting). This suggests that the intended behavior of FHD, reallocating representational capacity to improve storage efficiency while preserving fidelity, is robust to hyperparameter variations.

\begin{table*}[htbp]
\begin{center}
\setlength\tabcolsep{5pt}
\renewcommand{\arraystretch}{1.1}
\scalebox{0.67}{
\begin{tabular}{ c |cc cc cc cc cc cc}
\bottomrule \hline
\noalign{\smallskip}
Level & \multicolumn{2}{c}{Bike$_1$}& \multicolumn{2}{c}{Bike$_2$}& \multicolumn{2}{c}{Corgi}& \multicolumn{2}{c}{Yoga}& \multicolumn{2}{c}{Dance}& \multicolumn{2}{c}{Average}\\  
\hline\noalign{\smallskip}

1 
& 22.39 / 0.679 / \second{0.306} & 66
& \second{22.74} / \second{0.680} / \best{0.314} & 67
& \best{20.31} / 0.584 / \second{0.457} & 61.5
& \second{21.28} / \second{0.789} / \second{0.297} & 74.5
& \second{18.05} / \best{0.617} / 0.357 & 95.5
& \second{20.95} / 0.670 / \best{0.346} & 72.9 \\

2     
& \best{22.54} / \second{0.689} / \best{0.305} & \second{52.5}
& \best{22.76} / \best{0.684} / \second{0.315} & \second{54.5}
& \second{20.26} / \best{0.587} / 0.459 & \second{54}
& 20.95 / 0.788 / \best{0.296} & \second{68.8}
& 17.89 / \second{0.615} / \best{0.353} & \second{86.8}
& 20.88 / \second{0.673} / \best{0.346} & \second{63.3} \\

3   
& \second{22.51} / \best{0.691} / 0.309 & \best{48.8}
& 22.68 / \best{0.684} / \second{0.315} & \best{44}
& \best{20.31} / \second{0.586} / \best{0.452} & \best{50.7}
& \best{21.42} / \best{0.795} / \second{0.297} & \best{61.3}
& \best{18.09} / \best{0.617} / \second{0.355} & \best{81.8}
& \best{21.00} / \best{0.675} / \best{0.346} & \best{57.3} \\
\bottomrule \hline
\end{tabular}
}
\caption{\textbf{Ablation on the number of hierarchy levels in FHD on the $\textbf{SelfCap}_{\text{LR}}$ dataset}. “Level” indicates the number of anchor-point hierarchy levels, and we ablate this granularity by varying this count. Each block element of 4-performance denotes (PSNR(dB)$\uparrow$ / SSIM$\uparrow$ / LPIPS$\downarrow$ / Storage(MB)$\downarrow$).} 
\label{table:FHD_comparison}
\end{center}
\end{table*}

\begin{table}[h]
\centering
\renewcommand{\arraystretch}{1.15}
\setlength{\tabcolsep}{3pt}
\resizebox{\linewidth}{!}{
\begin{tabular}{c | c | c c c | c}
\hline
Hyperparameter & Value & PSNR $\uparrow$ & SSIM $\uparrow$ & LPIPS $\downarrow$ & Storage (MB) $\downarrow$\\
\hline
\multirow{3}{*}{\shortstack{$[\tau_1, \tau_2]$\\(Quantile Thresholds)\\ Eq. (2)}} 
& \textbf{[0.4, 0.7]} & 21.00 & 0.675 & 0.346 & \textbf{57.3} \\
& [0.8, 0.9] & 21.01 & 0.668 & 0.345 & 67.3 \\
& [0.1, 0.2] & 21.02 & 0.674 & 0.348 & 57.6 \\
\hline

\multirow{3}{*}{\shortstack{$[\lambda_0, \lambda_1, \lambda_2]$\\(Initial Level Importance) \\ Eq. (3)}} 
& \textbf{[1.0, 0.7, 0.6]} & 21.00 & 0.675 & 0.346 & \textbf{57.3} \\
& [1.0, 0.3, 0.3] & 20.93 & 0.670 & 0.346 & 60.43 \\
& [1.0, 1.5, 1.5] & 21.09 & 0.664 & 0.347 & 67.2 \\
\hline
\end{tabular}}
\caption{\textbf{Sensitivity analysis of FHD hyperparameters on the $\textbf{SelfCap}_{\text{LR}}$-S dataset}. Bold entries indicate our default setting. Each block element of 4-performance denotes (PSNR(dB)$\uparrow$ / SSIM$\uparrow$ / LPIPS$\downarrow$ / Storage(MB)$\downarrow$).}
\label{tab:hyperparam_ablation}
\end{table}

\begin{figure*}[htbp]
\centering
\includegraphics[scale=0.62]{fig/Suppl_Figure_10.pdf}
\caption{\textbf{FHD level-wise rendering comparison}.(a) Result trained without FHD, using all anchor-points and thus the largest anchor-point set. (b–d) Renderings that retain only Gaussians attached to Level 0–2 anchor-points, where higher levels progressively capture high-frequency and dynamic regions. (e) Combining Levels 0 and 1 reconstructs most static content with a small number of anchor-points. (f) Combining all levels merges the dynamic regions from higher levels into the final reconstruction.}
\label{fig:FHD_vis}
\end{figure*}

We qualitatively verify that variance-based leveling induces a frequency-aware hierarchy. In Fig.~\ref{fig:analyze}, the top row (a--c) presents level-specific renderings that selectively include only the anchor-points assigned to Levels 0--2 by VL. Rendering with only Level~0 primarily recovers global low-frequency scene structure (coarse shapes), whereas higher levels increasingly emphasize locally complex regions and fine details. The bottom row (d--f) reports the corresponding 2D FFT magnitudes, where the spectra progressively shift toward stronger high-frequency components from Level~0 to Level~2. This indicates that the VL-assigned hierarchy consistently reflects local frequency complexity and that FHD allocates higher-level representational capacity to regions requiring high-frequency refinement.

Fig.~\ref{fig:FHD_vis} further decomposes the contribution of each level under the three-level setting. Without FHD (a), anchors grow uniformly, resulting in the largest $\mathbf{A}^{\text{Key}}_n$. With FHD, Level~0 (b) captures global low-frequency structure but fails to fully recover fine textures and dynamic regions, while Level~1 (c) complements it by capturing mid-frequency details and moderate motions. Level~2 (d) concentrates on high-frequency textures and fast-moving regions with a small number of anchor-points, demonstrating that FHD reserves the highest level for the most demanding content. Combining Levels~0 and~1 (e) reconstructs most static content with substantially fewer anchor-points than (a), and aggregating all levels (f) restores the remaining dynamic and high-frequency content. This progression explains why finer level assignment improves both fidelity and storage efficiency.
\begin{figure}[htbp]
\centering
\includegraphics[scale=0.56]{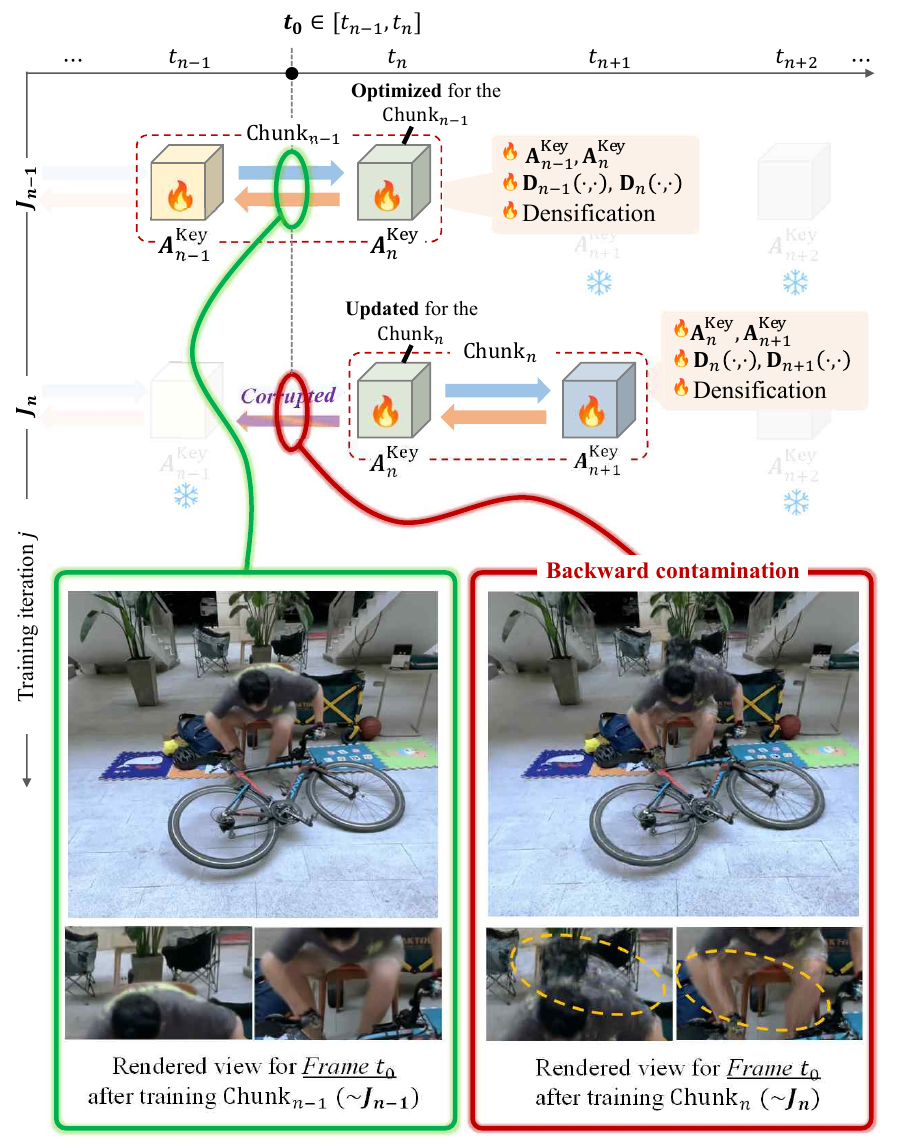}
\caption{\textbf{Backward contamination.} Example visualization and rendered patches illustrating the backward contamination issue in naïve chunk-wise training of bidirectional deformation.}
\label{fig:backward contamination}
\end{figure}

\subsection{Backward Contamination}
\label{Backward Contamination}
An illustration of the inter-chunk interference, referred to as \textit{backward contamination}, described in Sec.~\ref{sec:bidirectional deformation} is shown in Fig.~\ref{fig:backward contamination}. 
When bidirectional deformation is trained in a naïve chunk-wise manner, the result optimized for $\text{Chunk}_{n-1}$ during $J_{n-1}$ iterations can be corrupted by the subsequent updates of $\mathbf{A}^{\text{Key}}_n$ during $J_{n}$ iterations. 
The two rendered images correspond to the same time $t_0\in[t_{n-1}, t_n]$; the left image (green box) shows the result after completing only $J_{n-1}$ iterations of $\text{Chunk}_{n-1}$, while the right image (red box) shows the result after completing $J_{n}$.
As observed in the patch regions, the rendering after $J_{n-1}$ exhibits clean convergence in object areas. 
However, after training proceeds to $J_{n}$, newly grown anchor-points for optimizing $\text{Chunk}_{n}$ remain as \textit{ghost-like} residues because they are never trained in the backward direction for $\text{Chunk}_{n-1}$. 
Meanwhile, anchor-points that were crucial for $\text{Chunk}_{n-1}$ may be pruned, degrading the clearness of the object. This phenomenon is highlighted by the orange dotted ellipses in the rendered patches. 
Consequently, naïve chunk-based training inherently suffers backward contamination. 
In contrast, our PWD training and IFB training strategies (Sec.~\ref{sec:bidirectional deformation}) effectively eliminate such backward contamination.

\subsection{Temporal Profile Visualization}
\label{Temporal Profile Analysis}
\begin{figure}[htbp]
\centering
\includegraphics[scale=0.56]{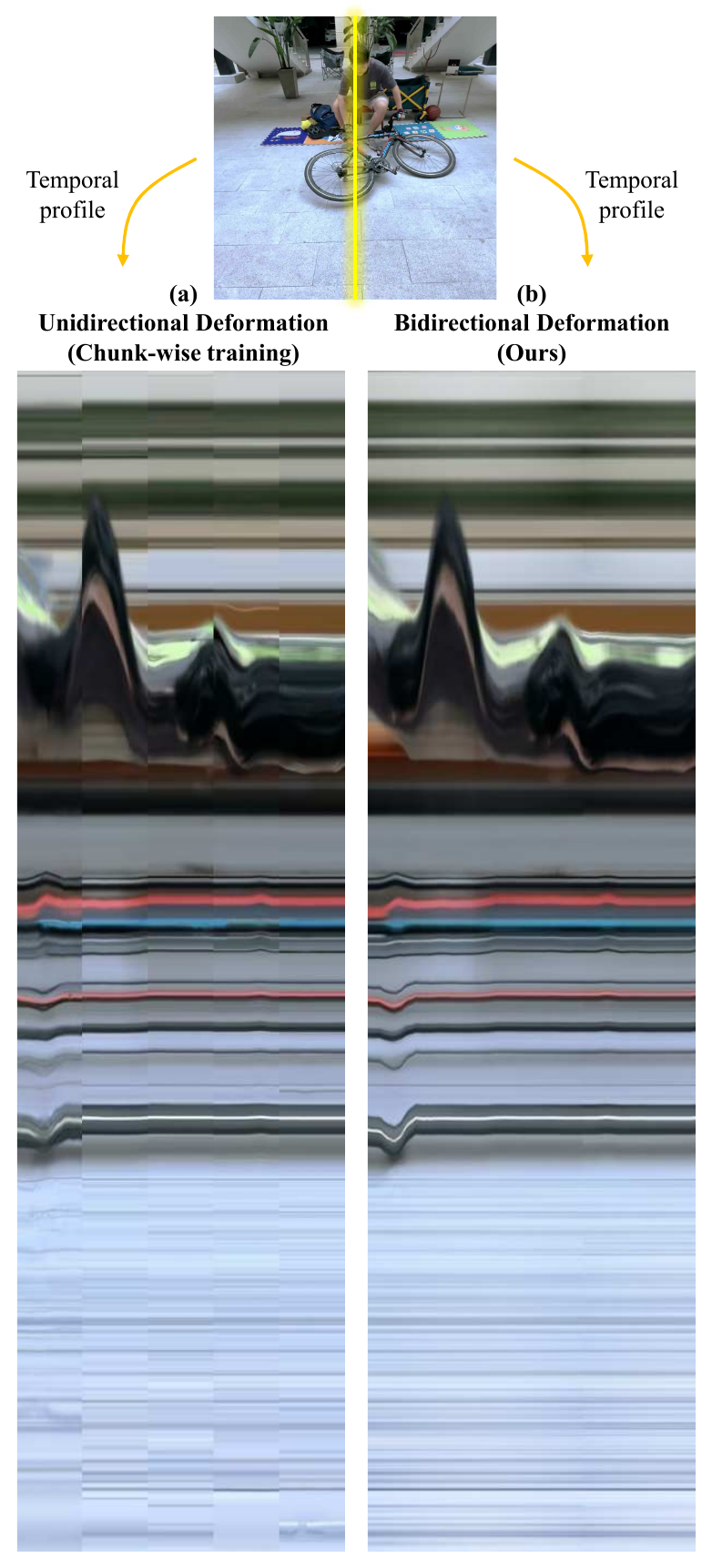}
\vspace{-1mm}
\caption{\textbf{Temporal profile visualization.} (a) Unidirectional deformation produces visibly distinguishable chunk boundaries, whereas (b) our bidirectional deformation yields smooth temporal continuity.}
\label{fig:temporal_profile}
\end{figure}

Fig.~\ref{fig:temporal_profile} visualizes the temporal flickering issue that arises in unidirectional deformation, using a temporal profile representation. 
The profile is constructed by accumulating a 1D scanline at the center of the rendered frame over time, producing a 2D image. 
Fig.~\ref{fig:temporal_profile}(a) shows the result of a chunk-wise trained unidirectional method, while Fig.~\ref{fig:temporal_profile}(b) presents the result of our bidirectional deformation. 
As seen in (a), visibly distinguishable discrete boundaries appear at every chunk transition, whereas in (b) such boundaries disappear, exhibiting smooth temporal continuity. 
This temporal discontinuity manifests as temporal flicker in actual rendered videos, significantly degrading perceptual quality.

\section{Limitations and Future Works}

While our method provides strong scalability and temporal consistency for long-range videos, it remains challenging to handle scenes in which the spatial extent becomes significantly larger or the spatial characteristics change over time. The GCA, which is introduced to maintain global temporal coherence, becomes less effective when the scene undergoes substantial spatial changes. 
In such cases, the GCA would need to be re-initialized (analogous to introducing a new large chunk), and temporal inconsistencies may appear around this transition.
To address spatial variation while preserving the temporal consistency of our approach, one promising direction is to incorporate spatial grids or frequency-aware representations, as explored in prior works~\cite{tancik2022block,zhang2025lookcloser}. 
These components could potentially be integrated with our chunk-wise \textit{on-demand} temporal loading strategy and the feature-variance–based frequency approximation used in MoRel. 
We envision extending this idea toward a unified framework that handles spatio-temporally large-scale motion more effectively, which we leave as an important avenue for future work.


\end{document}